%% file: arXiv.tex
\documentclass[10pt,twocolumn,letterpaper]{article}

\usepackage{cvpr}              
\makeatletter
\@namedef{ver@everyshi.sty}{}
\makeatother
\usepackage{tikz}

\usepackage{graphicx}
\usepackage{amsmath}
\usepackage{amssymb}
\usepackage{booktabs}
\usepackage{algorithm}
\usepackage{algorithmicx, algpseudocode}
\usepackage{nth}
\usepackage{color}
\usepackage{multirow}
\usepackage{arydshln}
\usepackage{threeparttable}
\usepackage[symbol]{footmisc}

\usepackage{titling}

\usepackage[utf8]{inputenc}
\usepackage{newunicodechar}
\newunicodechar{−}{-}     

\usepackage[pagebackref,breaklinks,colorlinks]{hyperref}

\usepackage{tikz}
\usepackage[accsupp]{axessibility}

\usepackage[capitalize]{cleveref}
\crefname{section}{Sec.}{Secs.}
\Crefname{section}{Section}{Sections}
\Crefname{table}{Table}{Tables}
\crefname{table}{Tab.}{Tabs.}

\begin{document}
\newcommand{\jy}[1]{\textcolor{blue}{\small [\textbf{JC}: #1]}}
\newcommand{\sy}[1]{\textcolor{orange}{\small [\textbf{SY}: #1]}}
\newcommand{\opt}[1]{\textcolor{teal}{\small [\textbf{OPT}: #1]}}
\newcommand{\rmv}[1]{\textcolor{red}{\small [#1]}}

\title{MEIL-NeRF: Memory-Efficient Incremental Learning\\ of Neural Radiance Fields}
\author{Jaeyoung Chung$^1$\qquad\quad Kanggeon Lee$^1$\qquad\quad Sungyong Baik$^2$\qquad\quad Kyoung Mu Lee$^1$\\
\\
$^1$Dept. of Electrical and Computer Engineering, ASRI, Seoul National University, Seoul, Korea\\
$^2$ Dept. of Data Science, Hanyang University, Seoul, Korea\\
{\tt\small robot0321@snu.ac.kr, dlrkdrjs97@snu.ac.kr, dsybaik@hanyang.ac.kr, kyoungmu@snu.ac.kr}\\
}
\date{}
\maketitle

\input{sections/0abstract}
\input{sections/1introduction}
\input{sections/2relatedworks}
\input{sections/3problemstatement}
\input{sections/4method}
\input{sections/5experiment}
\input{sections/6limitation7conclusion}
\pagebreak
\clearpage

\title{Supplementary Material \textit{for} \\ \vspace{2mm} MEIL-NeRF: Memory-Efficient Incremental Learning\\ of Neural Radiance Fields}
\maketitle

\setcounter{section}{0}
\setcounter{figure}{0}
\setcounter{table}{0}
\setcounter{equation}{0}

\renewcommand{\thetable}{S\arabic{table}}
\renewcommand{\thesection}{S\arabic{section}}
\renewcommand{\thefigure}{S\arabic{figure}}
\renewcommand{\theequation}{S\arabic{equation}}

\newcommand{\fakeref}[1]{\textcolor{red}{#1}}
\newcommand{\fakeeqref}[1]{(\textcolor{red}{#1})}
\newcommand{\fakecite}[1]{[\textcolor{green}{#1}]}

\input{sections/8suppleWithoutFakeref.tex}

\pagebreak
\clearpage

{\small
\bibliographystyle{ieee_fullname}
\bibliography{egbib}
}

\end{document}

%% file: sections/0abstract.tex
\begin{abstract}
Hinged on the representation power of neural networks, neural radiance fields (NeRF) have recently emerged as one of the promising and widely applicable methods for 3D object and scene representation.
However, NeRF faces challenges in practical applications, such as large-scale scenes and edge devices with a limited amount of memory, where data needs to be processed sequentially. 
Under such incremental learning scenarios, neural networks are known to suffer catastrophic forgetting: easily forgetting previously seen data after training with new data.
We observe that previous incremental learning algorithms are limited by either low performance or memory scalability issues.
As such, we develop a \textbf{M}emory-\textbf{E}fficient \textbf{I}ncremental \textbf{L}earning algorithm for \textbf{NeRF} (\textbf{MEIL-NeRF}).
MEIL-NeRF takes inspiration from NeRF itself in that a neural network can serve as a memory that provides the pixel RGB values, given rays as queries.
Upon the motivation, our framework learns which rays to query NeRF to extract previous pixel values.
The extracted pixel values are then used to train NeRF in a self-distillation manner to prevent catastrophic forgetting.
As a result, MEIL-NeRF demonstrates constant memory consumption and competitive performance.
\end{abstract}

%% file: sections/1introduction.tex
\section{Introduction} \label{section:introduction}
Representation and reconstruction of 3D objects and scenes are important computer vision and computer graphics tasks, with a broad range of applications, such as virtual reality~\cite{bruno20103d, gafni2021dynamic}, autonomous driving~\cite{ma2019accurate, janai2020computer}, and robotics~\cite{durrant2006simultaneous, bailey2006simultaneous, mur2015orb}.
Recently, neural radiance fields (NeRF)~\cite{mildenhall2020nerf} has brought substantial improvement by exploiting the representation power of neural networks.
They represent a static scene with an MLP network by querying spatial location and view direction along camera rays and integrating the output colors and densities using volume rendering techniques.
\begin{figure}[t]
    \includegraphics[width=1.0\columnwidth]{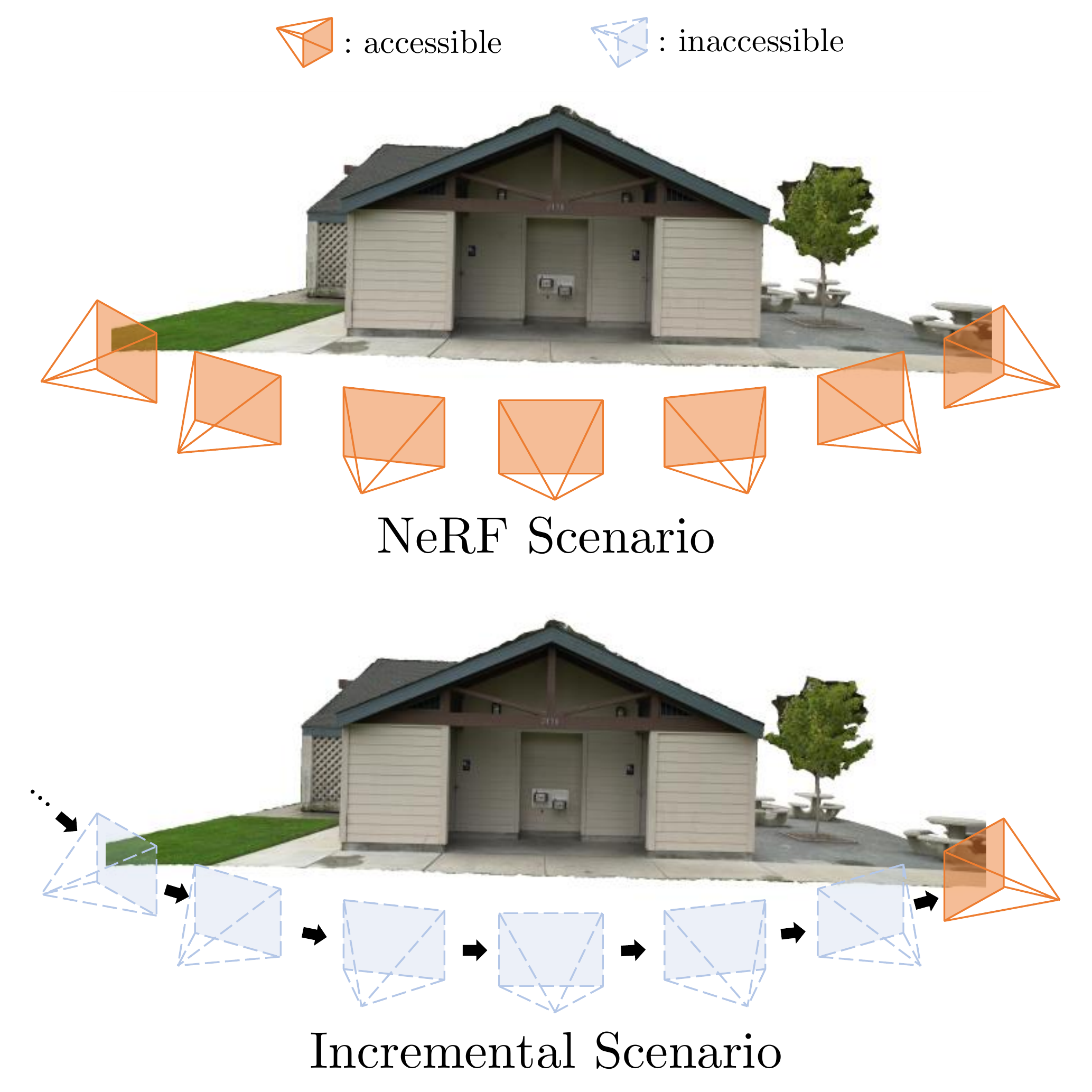}
    \vspace{-1.8em}
    \caption{\textbf{NeRF under standard scenario vs. incremental scenario.}
    Unlike vanilla NeRF where all viewpoints are always accessible, we consider incremental scenarios where only recent few viewpoints are accessible at each time.
    In the incremental scenarios, a set of images and corresponding poses are sequentially given and there is no or limited access to the past viewpoints.
    Such incremental scenario poses challenges in that NeRF will be susceptible to forgetting of images at previous viewpoints, degrading the overall representation performance.}
    \label{fig:fig1}
    \vspace{-1.4em}
\end{figure}
\begin{figure*}
\centering
\begin{subfigure}[b]{.49\linewidth}
\includegraphics[width=\columnwidth]{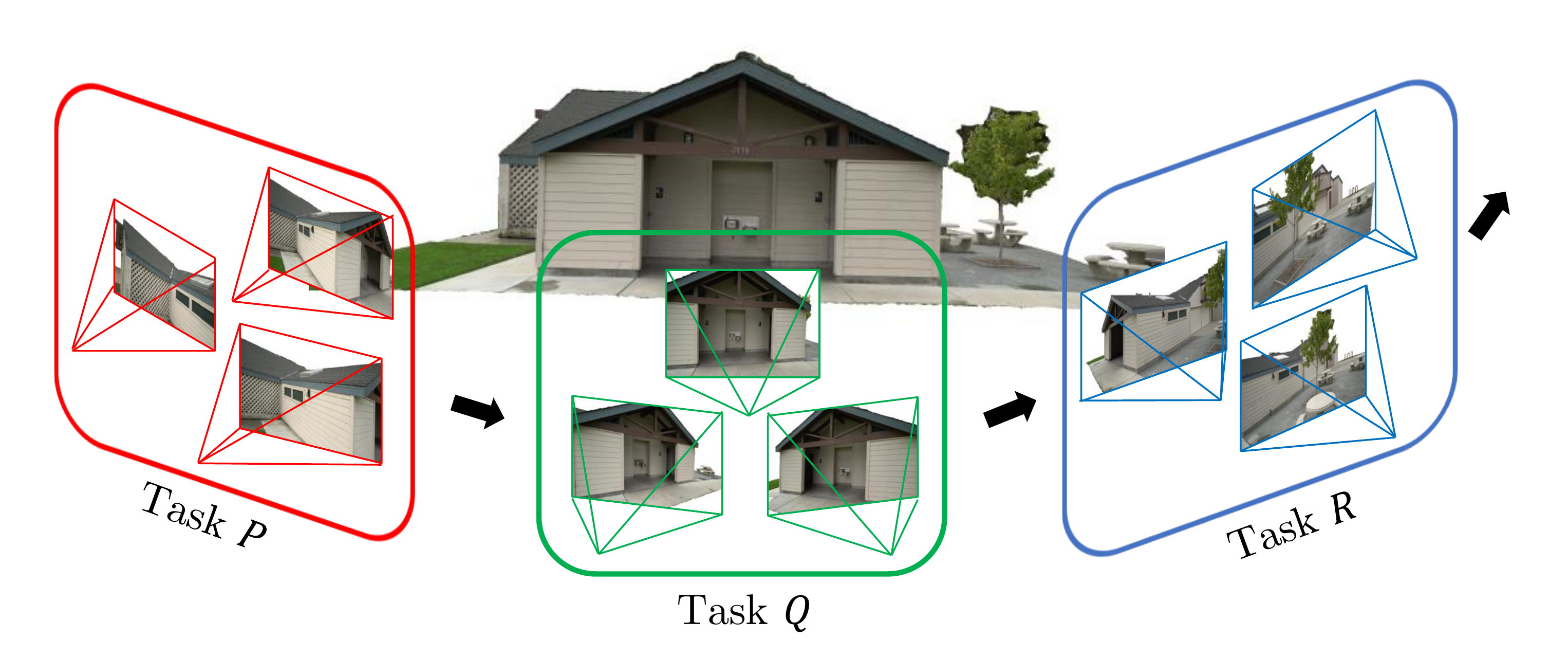}
\caption{An Example Data Sequence}\label{fig:fig2a}
\end{subfigure}
\begin{subfigure}[b]{.49\linewidth}
\includegraphics[width=\columnwidth]{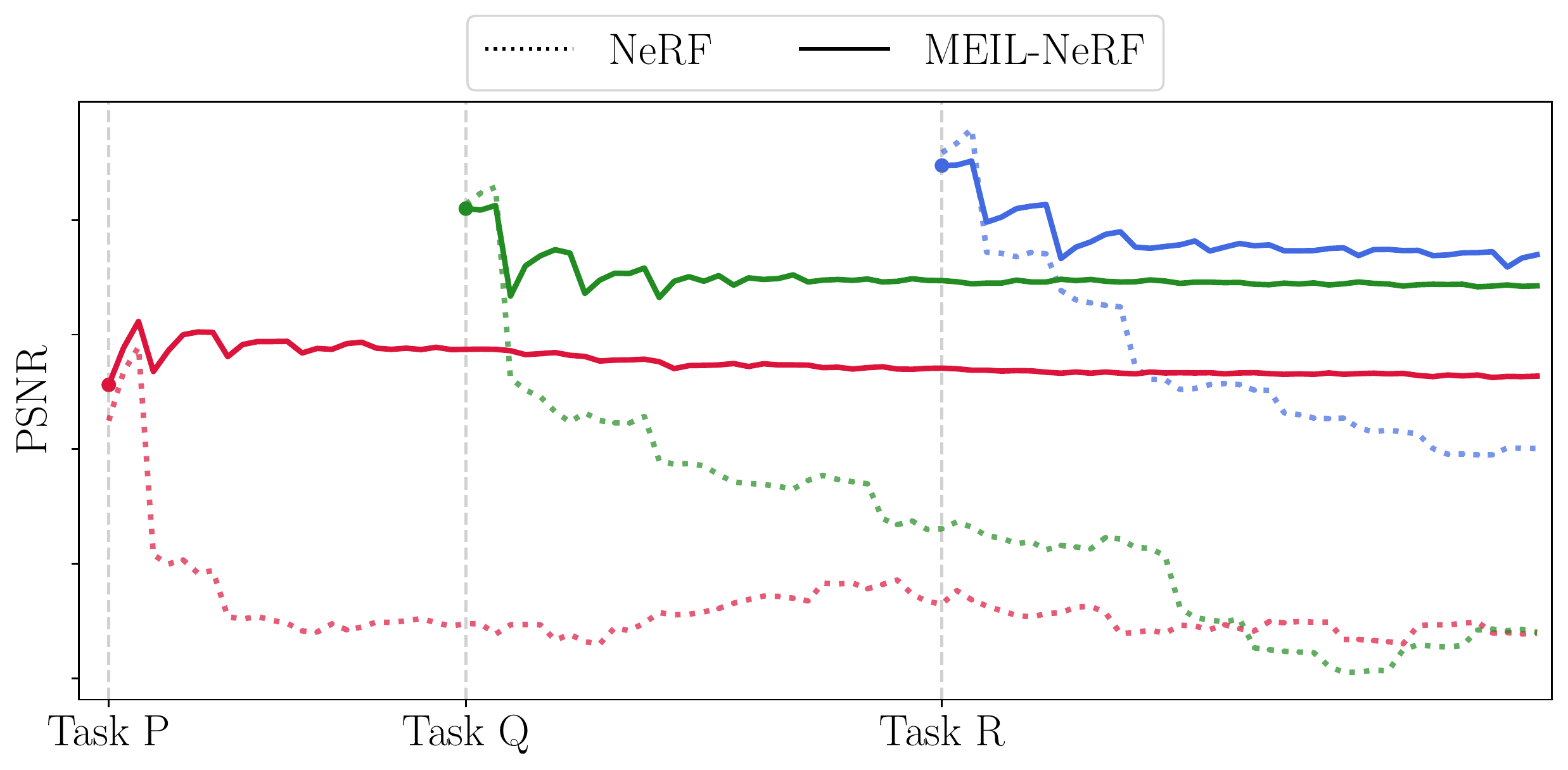}
\caption{NeRF and MEIL-NeRF in Incremental Scenario}\label{fig:fig2b}
\end{subfigure}
\vspace{-0.5em}
\caption{\textbf{Catastrophic Forgetting on NeRF in Incremental Scenario.} 
    We present an incremental scenario on NeRF and demonstrate the results.
    (a) shows an example data sequence, assuming that the NeRF network can only learn the data from the latest task.
    In this scenario, the performance of NeRF network drops rapidly, and the early tasks are completely ruined, as shown in (b).
    To alleviate catastrophic forgetting in the network, we propose a learning strategy based on self-distillation, which significantly slows down the forgetting speed.
}
\label{fig:fig2}
\vspace{-1.4em}
\end{figure*}
To achieve such outstanding performance, however, NeRF assumes access to all data (i.e., RGB values of a scene seen from all viewpoints) at once.
Such constraint hinders NeRF from being applied to practical applications---for example, large-scale scenes and edge devices with a limited amount of memory---where data needs to be processed sequentially.
In other words, NeRF will incrementally learn the scene, the partial information of which will be visible from few viewpoints at each time, as illustrated in Figure~\ref{fig:fig1}.

Under such incremental learning scenarios, neural networks are known to suffer catastrophic forgetting~\cite{french1999catastrophic}: old knowledge gets forgotten while learning new knowledge.
Various incremental learning algorithms have been developed to mitigate the adverse effects of catastrophic forgetting~\cite{de2021continual} for classification problems.
Among them, the method of saving a small portion of data in memory for replay has drawn attention for its simplicity and high performance.
A few works~\cite{sucar2021imap, zhu2022nice} have employed such replay method along with NeRF on simultaneous localization and mapping (SLAM).
Despite the simplicity and notable performance, such replay-based methods have a critical drawback to be utilized in practical applications: scalability issues due to monotonically increasing memory.

Considering the lack of studies on incremental learning problems with NeRF, we first introduce a new benchmark, where the camera does not revisit the previously seen part of the scene, which becomes no longer accessible afterwards (Figure~\ref{fig:fig2a}).
Then, we train NeRF with several representative incremental learning algorithms to evaluate its effectiveness in the context of 3D representation.
We observe that they suffer low performance or memory scalability issues.
The low performance is mainly due to the absence of the access to previous data, whereas the memory scalability issue is due to the need for increasing memory to store previous data.

To balance the trade-offs, we introduce a \textbf{M}emory-\textbf{E}fficient \textbf{I}ncremental \textbf{L}earning of \textbf{NeRF} (\textbf{MEIL-NeRF}).
In order to prevent catastrophic forgetting without increasing memory, we turn our attention to a neural network itself.
In NeRF, a neural network produces the pixel RGB values, when rays or viewpoints are given as input.
Thus, we consider a neural network as a memory storage for the pixel RGB values of the scene.
The proposed perspective raises a question: which rays should we give to NeRF such that previously seen pixel RGB values of the scene will be retrieved?
Only when rays are directed at the scene, we will be able to extract the pixel RGB values of the scene from the network.
In this work, we answer the question by introducing another small network, named Ray Generator Network (RGN), that is trained to produce the previously seen rays directed towards the scene.
Then, we feed the produced rays into NeRF to obtain previous RGB values of the scene, which are used to train NeRF in a self-distillation~\cite{zhang2019your} manner to prevent forgetting of previous RGB values.

The experimental results demonstrate that our proposed framework greatly reduces the adverse effects of catastrophic forgetting in NeRF without increasing memory, as shown in Figure~\ref{fig:fig2b}.
The memory-efficient incremental learning algorithm is made feasible by considering NeRF as memory storage and using RGN to remember which rays to feed NeRF to extract previous RGB values.

%% file: sections/2relatedworks.tex
\begin{figure*}[t]
    \includegraphics[width=1.0\textwidth]{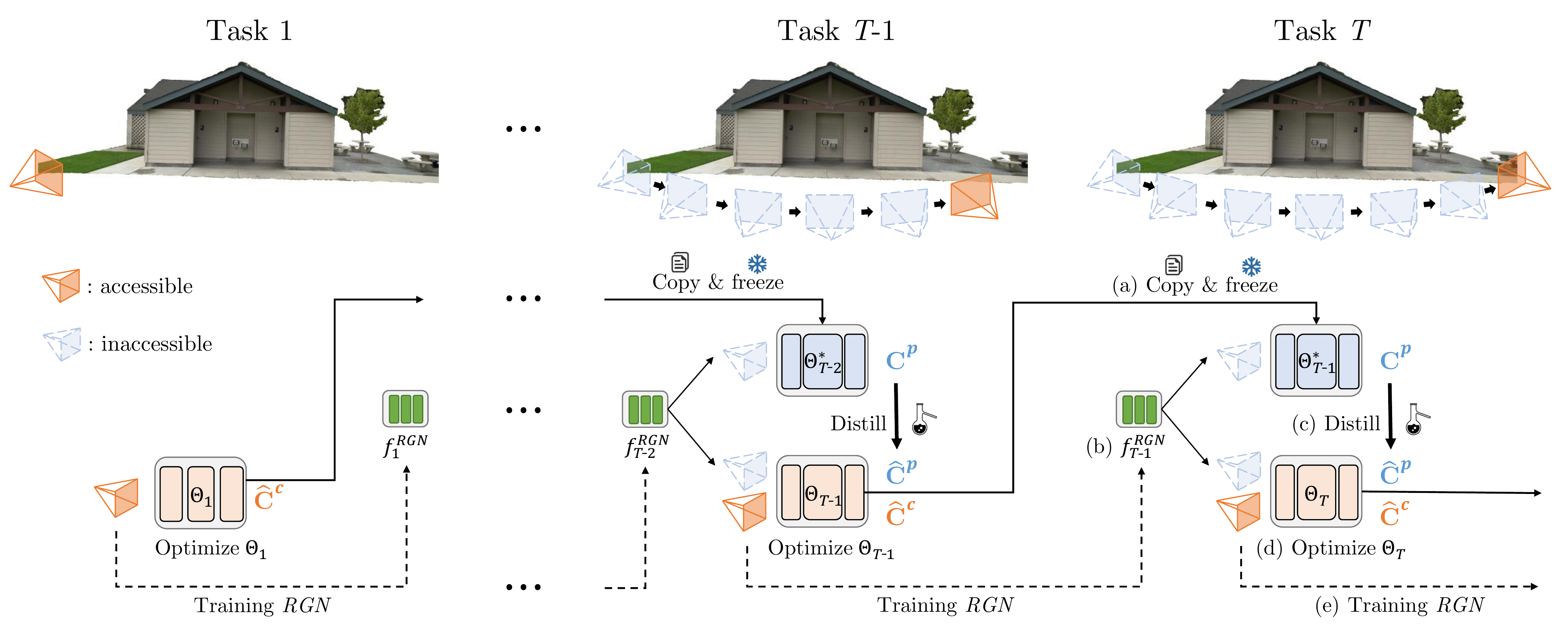}
    \caption{\textbf{Overall Pipeline of Proposed Method.} 
    We present the learning details of the proposed method.
    (a) Before starting each task, we copy and freeze the previous parameters for distillation.
    (b) We create rays for distillation using \textit{Ray Generator Network} (RGN) that remembers which rays are important.
    (c) We retrieve past color information by querying sample points along the generated rays from the saved parameters which is treated as memory storage.
    (d) In each task, the network learns current task data(orange) and the distilled past data(blue) at the same time. 
    (e) After the training of each task ends, we update RGN with current task rays and distilled past task rays.
    }
    \label{fig:fig3}
    \vspace{-1.4em}
\end{figure*}

\section{Related Works}
\noindent
\textbf{Neural Radiance Fields.}
There has been several attempts in implicitly expressing 3D geometry either by learning signed distance function~\cite{park2019deepsdf}, occupancy probability~\cite{mescheder2019occupancy}, or color information at spatial location~\cite{sitzmann2020implicit,mildenhall2020nerf}.
Among the attempts, neural radiance fields (NeRF)~\cite{mildenhall2020nerf} has emerged as one of promising methods for its high reconstruction quality, spurring a large number of research works on improvements~\cite{martin2021nerf, tancik2020fourier, barron2021mip, barron2022mip, verbin2022ref, kellnhofer2021neural, yang2022banmo, yu2021pixelnerf, liu2020nsvf, lin2021barf, tancik2021learned, hedman2021baking, yu2021plenoxels, wang2021ibrnet, chen2021mvsnerf, muller2022instant, reiser2021kilonerf} and a wide range of applications~\cite{xie2022neural}: controllable human avatar~\cite{peng2021animatable, chen2021animatable, lombardi2021mixture}, realistic game~\cite{hao2021gancraft}, robotics~\cite{sucar2021imap, zhu2022nice, ichnowski2021dex, oechsle2021unisurf}, 3D-aware image generation~\cite{chan2021pi, chan2022efficient}, and data compression~\cite{dupont2021coin}.
Given images and the corresponding camera poses, NeRF casts camera rays, queries sample points and view direction, and finally synthesizes an image by accumulating the output colors and densities along the camera rays.
Since NeRF assumes access to all data (i.e., viewpoints and the corresponding scene colors) for training, NeRF faces challenges in practical applications (e.g., edge devices with a limited amount of memory), where such assumption does not hold.
Consequently, NeRF is required to learn the scene with the online stream of data without access to previously observed data.

\noindent
\textbf{Incremental learning.}  
Under the incremental learning scenario as described above, neural networks are known to forget the previously learned knowledge while learning new knowledge, which is referred to as catastrophic forgetting~\cite{french1999catastrophic}.
There has been many studies to mitigate catastrophic forgetting for classification tasks.
Incremental learning algorithms can be divided into three methodologies: regularization, parameter isolation, and replay~\cite{de2021continual}.
Regularization-based methods aim to retain the learned mapping for past data (old knowledge) while learning a mapping for new data (new knowledge).
To do so, several works~\cite{kirkpatrick2017overcoming, zenke2017continual, aljundi2018memory, chaudhry2018riemannian} try to find which parameters are important for old knowledge and regularize the changes in those important parameters while learning new knowledge.
In contrast, other works aim to regularize the changes in outputs for previously learned classes~\cite{Li2018learning, dhar2019learning} while learning new classes in the context of classification. 
Meanwhile, parameter isolation methods~\cite{mallya2018packnet, mallya2018piggyback, hung2019compacting, rosenfeld2018incremental} attempt to learn a separate sub-network for each type of data (i.e., task).
Last but not least, replay-based methods aim to preserve old knowledge by storing the selected samples of past data, which are used together with new data during training. 

\noindent
\textbf{Incremental Learning for Neural Radiance Fields.}
There has been very few studies on trying to incorporate NeRF with incremental learning tailored for specific applications, such as simultaneous localization and mapping (SLAM).
iMAP~\cite{sucar2021imap} continuously select and save few data at each keyframe.
On the other hand, NICE-SLAM~\cite{zhu2022nice} and NeRF-SLAM~\cite{rosinol2022nerf} employ hierarchical volumetric structures by reserving multi-scale spatial grid locations with feature representation and different NeRF, respectively.
However, these methods mostly suffer the memory scalability issues as the required amount of memory increases (either for model or data) with the scale of space.

In this work, we observe that prior works either suffer low performance or memory inefficiency.
To balance between the trade-offs, we propose a new perspective that NeRF can act as a storage: given appropriate rays directed at the scene, it gives the corresponding RGB values. 
Thus, if a framework learns which rays are directed at the scene, past data (RGB values seen from past viewpoints) can be easily retrieved by feeding the rays into NeRF.
Then, NeRF can be trained on new data along with retrieved past data to prevent forgetting.
Upon the motivation, we introduce another network, which learns which rays are relevant for the given scene.
As a result, our framework alleviates catastrophic forgetting for NeRF with constant memory under incremental learning scenarios.

%% file: sections/3problemstatement.tex
\section{Problem Statement} \label{section:problemstatement}
Before delving into details of the proposed framework, we start with defining the problem statement of neural radiance fields (NeRF) under incremental learning scenarios.
In particular, we consider task incremental learning scenarios~\cite{Li2018learning,de2021continual}, where a batch of data (i.e., task) comes in sequentially.
In the context of NeRF, each task $\mathcal{T}$ is constructed with $\mathcal{N}$ number of images and the corresponding camera poses, where $\mathcal{N} > 1$ to ensure that geometric information can be obtained at each task.
Specifically, $t$-th task $\mathcal{T}_t$ is made of a set of paired data $(\mathcal{X}^{(t)}, \mathcal{Y}^{(t)})$, where $\mathcal{X}^{(t)}$ is a set of rays emitted from the given $\mathcal{N}$ camera views and $\mathcal{Y}^{(t)}$ is a set of the corresponding RGB colors.
To impose incremental learning, only the latest task is available while previous tasks become inaccessible.
Overall, given the formulation, the objective of incremental learning is to find the optimal parameters $\Theta_\text{opt}$ that prevent catastrophic forgetting and perform well across all seen tasks:
\begin{align}
    \Theta_\text{opt} = \min_{\Theta}\sum_{t=1}^{T}\mathbb{E}_{(\mathcal{X},\mathcal{Y}\sim\mathcal{D})}\left[\mathcal{L}(\mathcal{F}(\mathcal{X}^{(t)};\Theta),\mathcal{Y}^{(t)})\right],
\end{align}
where $T$ is the number of seen task; $\mathcal{L}$ is a loss function; $\mathcal{F}$ is a forward mapping (consisting of NeRF network and volume rendering) from $\mathcal{X}^{(t)}$ to $\mathcal{Y}^{(t)}$; and $\Theta$ denotes the parameters of the network.

%% file: sections/4method.tex
\section{Method}
In this section, we start with a formulation of neural radiance fields (NeRF) in Section~\ref{section:method_preliminary}.
Then, we delineate our proposed framework, MEIL-NeRF, which prevents the catastrophic forgetting by using past task information retrieved from NeRF itself (Section~\ref{section:method_selfdistill}) with generated rays from our newly introduced ray generator network (Section~\ref{section:method_RGN}). 
The overall pipeline is illustrated in Figure~\ref{fig:fig3}.

\subsection{Preliminaries}\label{section:method_preliminary}
NeRF~\cite{mildenhall2020nerf} aims to represent a scene with an MLP network that takes 3D coordinates $\mathbf{p}$ and view direction $\mathbf{r}_d$ as input and produces RGB color $\mathbf{c}$ and volume density $\sigma$ as output.
The corresponding pixel color $\mathbf{C}$ of the given camera ray $\mathbf{r}$ is estimated by volume rendering~\cite{kutulakos1999theory}, summing with quadrature rule~\cite{max1995optical} over sampled 3D points along a view direction $\mathbf{r}_d$ of given camera ray $\mathbf{r}=(\mathbf{r}_o,\mathbf{r}_d)$, where each sampled point is obtained by $\mathbf{p}_i = \mathbf{r}_o + z_i*\mathbf{r}_d$; $r_o$ is the coordinate of a camera; and $z_i$ is the distance from a camera to the sampled point $\mathbf{p}_i$.
Thus, the estimated pixel color $\hat{\mathbf{C}}$ of the ray $\mathbf{r}$ is obtained by 
\begin{align}
    \hat{\mathbf{C}} = \sum_{i=1}^{P} \left(\prod_{j=1}^{i-1}(1-\alpha_j) \alpha_i \mathbf{c}_i\right),
\end{align}
where $\alpha_i=1-\exp(-\sigma_i(z_{i+1}-z_i))$ and $P$ is the number of sampled points.
To avoid notation clutter, we refer to the forward mapping from the ray to the pixel color as $\hat{\mathbf{C}} = \mathcal{F}(\mathbf{r};\Theta)$, where $\Theta$ is the parameters of the MLP network.

\input{sections/algorithm_main}

\subsection{MEIL-NeRF} \label{section:method_selfdistill}
The core idea of the proposed framework is to retrieve the pixel colors of past camera rays by querying the network with the past rays.
In this section, we describe how the proposed framework uses the retrieved pixel colors of past camera rays to prevent catastrophic forgetting, whereas Section~\ref{section:method_RGN} discusses how the past rays are generated.
To ensure that accurate past task information is retrieved from the network, we copy and free the parameters of the network before training the network on the current $T$-th task.
Since the network has been trained on $T-1$ previous tasks, we use $\Theta^*_{T-1}$ to denote the frozen parameters.
At each training iteration for $T$-th task, we obtain the pixel colors of past camera rays from the network with $\Theta^*_{T\text{-}1}$ by feeding it with the past rays $\mathbf{r}^p$:
\begin{align}
    \mathbf{C}^p = \mathcal{F}(\mathbf{r}^p{};\Theta^{*}_{{T}\text{-}1}). \label{eq:method_distill}
\end{align}
Treating $\mathbf{C}^p$ as pseudo ground truth, we optimize $\Theta_{T}$ to remember past tasks in a self-distillation~\cite{zhang2019your} manner.
Thus, the network is trained to learn the current task ${(\mathbf{r}^c, \mathbf{C}^c)}$ and past tasks $(\mathbf{r}^p, \mathbf{C}^p)$: 
\begin{align}
    \mathcal{L} &= \frac{1}{m_c} \sum_{\mathbf{r}^c}{\begin{Vmatrix} \hat{\mathbf{C}}^c - \mathbf{C}^c \end{Vmatrix}^2}+\frac{\lambda_p}{m_p} \sum_{\mathbf{r}^p}{\rho \begin{pmatrix}\hat{\mathbf{C}}^p - \mathbf{C}^p \end{pmatrix}}, \label{eq:method_loss}
\end{align}
where $\rho(x)=\sqrt{x^2+\epsilon^2}$ is a Charbonnier penalty function; $\lambda_p$ is a hyperparameter that controls the trade-off between learning current and past tasks; $m_c$ and $m_p$ are the batch size of the current and past rays; and $\hat{\mathbf{C}}^c$ and $\hat{\mathbf{C}}^p$ are estimated colors by the network that is being trained:
\begin{align}
    [\hat{\mathbf{C}}^c, \hat{\mathbf{C}}^p] = \mathcal{F}\left([\mathbf{r}^c, \mathbf{r}^p]{};\Theta_{{T}}\right). \label{eq:method_forward}
\end{align}
While we use L2 loss for learning current data as in standard NeRF training~\cite{mildenhall2020nerf}, we adopt Charbonnier penalty function~\cite{sun2010secrets} for learning past data.
Since Charbonnier is a differentiable variant of L1 distance, which better helps preserve past data without blurring that may be caused if L2 loss is employed otherwise.
When learning each task, we set $\lambda_p$ to be small at the beginning and then gradually increase it.
Such scheduling facilitates training by making the network focus on learning current task first and then preventing catastrophic forgetting. 

\input{sections/table_5exp_dataset_quan}

\subsection{Ray Generator Network} \label{section:method_RGN}
Thus, the na\"ive random generation of rays will most likely not point at the scene of interest, failing to provide data of past tasks.
Another possible alternative is to store the camera rays regarding past tasks, but it has the memory scalability issues.
Instead, we introduce a ray generator network (RGN), which is trained to produce the past camera rays that are directed to the scene of interest.
RGN is implemented by a small MLP network $f^{\text{RGN}}$ that learns to map from a real number $x\in[0,1]$ to the origin $\mathbf{r}_o^*$ and the direction $\mathbf{r}_d^*$ of a ray of unit length:
\begin{align}
    \left ( \mathbf{r}_{o}^*, \mathbf{r}_{d}^* \right ) = f^{\text{RGN}}\left(x\right ). \label{eq:method_rgn_generate}
\end{align}
Since we have $\mathcal{N}$ number of cameras and their corresponding principal rays for each task, the $({T}-1)\mathcal{N}$ equally spaced numbers $\mathbf{x}^{{T}\text{-}1} = [0, \frac{1}{(({T}\text{-}1)\mathcal{N}\text{-}1)}, \frac{2}{(({T}\text{-}1)\mathcal{N}\text{-}1)}, \cdots, 1]$ are assigned to the corresponding past principal rays, where $x=0$ and $x=1$ correspond to the initial and latest principal rays, respectively.
Centered around the generated principal rays $\left ( \mathbf{r}_{o}^*, \mathbf{r}_{d}^* \right )$, we use camera intrinsics to create $m_p$ non-principal rays $\left ( \mathbf{r}_{o}, \mathbf{r}_{d} \right )$ around the principal rays but with the same origin $\mathbf{r}_o = \mathbf{r}_o^*$:
\begin{align}
    \mathbf{r}_{d} = \frac{\text{f}\cdot\mathbf{r}_{d}^* + s\cdot\cos(u_{\theta})\cdot\mathbf{p}_1 + s\cdot\sin(u_{\theta})\cdot\mathbf{p}_2}{\lvert\text{f}\cdot\mathbf{r}_{d}^* + s\cdot\cos(u_{\theta})\cdot\mathbf{p}_1 + s\cdot\sin(u_{\theta})\cdot\mathbf{p}_2\rvert}, \label{eq:method_rgn_intrinsic}
\end{align}
where $\mathbf{p}_1, \mathbf{p}_2$ are unit vectors perpendicular to the principal ray obtained by the Gram-Schmidt process; $f$ is a focal length; and $s, u_\theta$ are random variables with uniform distributions as $s\sim\mathcal{U}[0,\sqrt{W^2+H^2}/2]$ and $u_\theta\sim\mathcal{U}[0,2\pi]$. 
Such formulation enables us to create a substantial amount of rays while reducing the learning complexity of RGN.
Overall, past rays can be retrieved from RGN by feeding it with $x\sim\mathcal{U}[0,1]$, along with non-principal rays by Equation~\eqref{eq:method_rgn_intrinsic}.

Since new rays likewise come in sequentially, we perform incremental learning of RGN via self-distillation. 
At the end of each task, we update RGN with the current principal rays $(\mathbf{r}^*_o, \mathbf{r}^*_d)_{{T}}$ and the past principal rays $(\mathbf{r}^*_{o}, \mathbf{r}^*_{d})_{1:{T}\text{-}1}$ that are retrieved from RGN conditioned on $\mathbf{x}^{{T}\text{-}1}$.
Then, we train RGN to map ${T}{\cdot}\mathcal{N}$ equally spaced numbers $\mathbf{x}^{{T}} \in [0,1]$ to $(\mathbf{r}^*_{o}, \mathbf{r}^*_{d})_{1:{T}}$ via MSE loss:
\begin{align}
    \mathcal{L}_{\text{RGN}} = \begin{Vmatrix} (\mathbf{r}^*_o,\mathbf{r}^*_d)_{1:{T}} - f^{\text{RGN}}(\mathbf{x}^{{T}}) \end{Vmatrix}^2. \label{eq:method_rgn_loss}
\end{align}

%% file: sections/algorithm_main.tex
\begin{algorithm}[t]
\caption{MEIL-NeRF}\label{alg:MEIL}
\begin{algorithmic}[1]
\Require{$\Theta_{T\text{-}1}$} 
\Comment{Parameters trained on previous tasks}
\State Copy and freeze as $\Theta_{{T}\text{-}1}^*$  
\For{iteration}
\State Sample current data $(\textbf{r}^c, \mathbf{C}^c)$
\If{$T>1$}
    \State Generate past rays $\textbf{r}^p$
    \Comment{Eq.~\ref{eq:method_rgn_generate},~\ref{eq:method_rgn_intrinsic}}
    \State Obtain past data $\left(\textbf{r}^p, \mathcal{F}(\textbf{r}^p;\Theta_{{T}\text{-}1}^*)\right)$
    \Comment{Eq.~\ref{eq:method_distill}}
\EndIf

\State Update NeRF network $\Theta_{{T}}$
\Comment{Eq.~\ref{eq:method_forward},~\ref{eq:method_loss}}
\EndFor
\State Update ray generator network $f^{\text{RGN}}$  \Comment{Eq.~\ref{eq:method_rgn_loss}}
\end{algorithmic}
\end{algorithm}

%% file: sections/table_5exp_dataset_quan.tex
\begin{table*}[t]
\centering
\caption{Quantitative results on the datasets. The best and second best performance are \textbf{highlighted} and \underline{underlined}, respectively.}
\vspace{-0.8 em}
\label{tab:dataset_quan}
\resizebox{\textwidth}{!}{%
\begin{threeparttable}
\begin{tabular}{|cc|ccccc|cccccc|ccc|}
\hline
\multicolumn{2}{|c|}{\multirow{2}{*}{}} &
  \multicolumn{5}{c|}{\textit{\textbf{Tanks and Temples (Custom)}}} &
  \multicolumn{6}{c|}{\textit{\textbf{Replica (Custom)}}} &
  \multicolumn{3}{c|}{\textit{\textbf{TUM-RGBD (Custom)}}} \\ \cline{3-16} 
\multicolumn{2}{|c|}{} &
  Barn &
  Caterpillar &
  Family &
  Ignatius &
  Truck &
  Office0 &
  Office3 &
  Office4 &
  Room0 &
  Room1 &
  Room2 &
  Seq0 &
  Seq1 &
  Seq2 \\ \hline
\multicolumn{1}{|c|}{\multirow{7}{*}{\rotatebox[origin=c]{90}{PSNR (dB)}}} &
  NeRF-Incre &
  14.17 &
  15.90 &
  23.38 &
  17.00 &
  16.57 &
  24.44 &
  16.28 &
  19.55 &
  19.18 &
  17.44 &
  19.10 &
  13.15 &
  13.72 &
  12.56 \\ 
\multicolumn{1}{|c|}{} &
  EWC~\cite{kirkpatrick2017overcoming} &
  14.70 &
  12.35 &
  19.66 &
  13.45 & 
  16.80 &
  24.01 &
  16.78 &
  18.31 &
  19.25 &
  17.88 &
  19.30 &
  13.60 &
  14.25 &
  12.56 \\
\multicolumn{1}{|c|}{} &
  PackNet~\cite{mallya2018packnet} &
  15.67 &
  12.77 &
  22.91 &
  19.99 &
  16.76 &
  28.68 &
  20.36 &
  24.54 &
  19.28 &
  24.53 &
  22.55 &
  17.66 &
  17.04 &
  17.46 \\
\multicolumn{1}{|c|}{} &
  iMAP\tnote{\dag}~\cite{sucar2021imap} &
  \underline{20.70} &
  \underline{21.36} &
  \underline{28.57} &
  \underline{24.05} &
  \underline{22.65} &
  \underline{33.82} &
  \underline{26.28} &
  \underline{31.57} &
  \underline{25.07} &
  \underline{28.14} &
  \underline{27.42} &
  \underline{21.89} &
  \underline{21.52} &
  \underline{21.44} \\
\multicolumn{1}{|c|}{} &
  MEIL-NeRF (ours) &
  \textbf{22.14} &
  \textbf{23.41} &
  \textbf{29.95} &
  \textbf{25.03} &
  \textbf{24.31} &
  \textbf{34.35} &
  \textbf{26.52} &
  \textbf{32.72} &
  \textbf{25.73} &
  \textbf{29.47} &
  \textbf{28.23} &
  \textbf{24.12} &
  \textbf{23.78} &
  \textbf{23.59} \\ \cdashline{2-16}[1pt/1.5pt]
\multicolumn{1}{|c|}{} &
  iMAP*~\cite{sucar2021imap} &
  {22.05} &
  {21.99} &
  {30.88} &
  {24.85} &
  {23.79} &
  {35.98} &
  {28.97} &
  {34.07} &
  {26.05} &
  {30.16} &
  {29.71} &
  {23.93} & 
  {23.61} &
  {23.30} \\
\multicolumn{1}{|c|}{} &
  NeRF-Joint~\cite{mildenhall2020nerf} &
  24.51 &
  25.91 &
  33.89 &
  26.59 &
  27.12 &
  39.99 &
  32.85 &
  37.67 &
  29.35 &
  34.36 &
  28.79 &
  25.91 &
  25.86 &
  25.96 \\ \hline
\multicolumn{1}{|c|}{\multirow{7}{*}{\rotatebox[origin=c]{90}{MS-SSIM}}} &
  NeRF-Incre &
  0.394 &
  0.657 &
  0.890 &
  0.747 &
  0.649 &
  0.679 &
  0.508 &
  0.649 &
  0.552 &
  0.558 &
  0.499 &
  0.345 &
  0.385 &
  0.299 \\
\multicolumn{1}{|c|}{} &
  EWC &
  0.405 &
  0.531 &
  0.755 &
  0.514 &
  0.657 &
  0.663 &
  0.526 &
  0.629 &
  0.558 &
  0.515 &
  0.504 &
  0.334 &
  0.382 &
  0.281 \\
\multicolumn{1}{|c|}{} &
  PackNet &
  0.423 &
  0.414 &
  0.877 &
  0.761 &
  0.549 &
  0.775 &
  0.558 &
  0.716 &
  0.487 &
  0.677 &
  0.631 &
  0.517 &
  0.480 &
  0.466 \\
\multicolumn{1}{|c|}{} &
  iMAP\tnote{\dag} &
  \underline{0.756} &
  \underline{0.874} &
  \underline{0.962} &
  \underline{0.940} &
  \underline{0.890} &
  \textbf{0.948} &
  \underline{0.895} &
  \underline{0.948} &
  \underline{0.773} &
  \underline{0.891} &
  \underline{0.875} &
  \underline{0.813} &
  \underline{0.782} &
  \underline{0.774} \\
\multicolumn{1}{|c|}{} &
  MEIL-NeRF (ours) &
  \textbf{0.836} &
  \textbf{0.909} &
  \textbf{0.972} &
  \textbf{0.953} &
  \textbf{0.925} &
  \textbf{0.948} &
  \textbf{0.908} &
  \textbf{0.960} &
  \textbf{0.831} &
  \textbf{0.914} &
  \textbf{0.896} &
  \textbf{0.882} &
  \textbf{0.864} &
  \textbf{0.858} \\ \cdashline{2-16}[1pt/1.5pt]
\multicolumn{1}{|c|}{} &
  iMAP*\tnote{\dag} &
  {0.838} &
  {0.841} &
  {0.981} &
  {0.952} &
  {0.917} &
  {0.970} &
  {0.934} &
  {0.968} &
  {0.828} &
  {0.919} &
  {0.917} &
  {0.878} &
  {0.853} &
  {0.835} \\
\multicolumn{1}{|c|}{} &
  NeRF-Joint &
  0.907 &
  0.944 &
  0.992 &
  0.966 &
  0.959 &
  0.987 &
  0.972 &
  0.984 &
  0.910 &
  0.972 &
  0.904 &
  0.919 &
  0.910 &
  0.911 \\ \hline
\end{tabular}%
\begin{tablenotes}
\small\item[\dag] reproduced to have similar memory usage to ours 
\small\item[*] reproduced to have similar performance to ours without restricting the memory usage
\end{tablenotes}
\end{threeparttable}
}
\vspace{-1.4 em}
\end{table*}

%% file: sections/5experiment.tex
\section{Experiment}  
We describe the experiment settings in Section~\ref{section:exp_expsettings}; dataset construction for incremental learning in Section~\ref{section:exp_dataset}; and report the experiment results and ablation studies in Section~\ref{section:exp_results} and \ref{section:exp_ablation}, respectively.
\begin{figure}[t]
    \includegraphics[width=1.0\columnwidth]{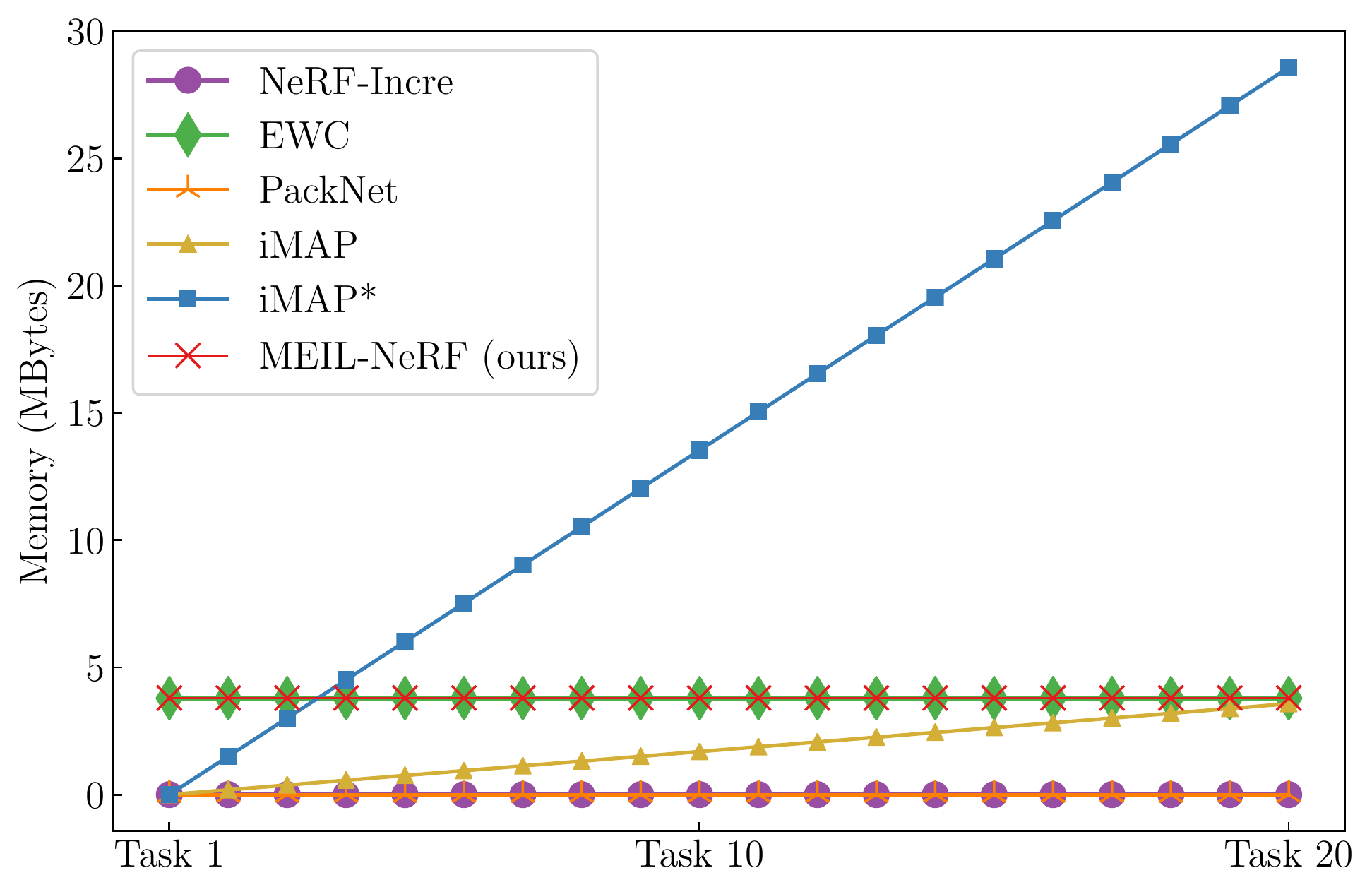}
    \vspace{-1.8em}
    \caption{\textbf{Additional Memory Usage.} 
    iMAP and iMAP* increases memory storage as it saves the ground truth data steadily.
    PackNet does not use additional memory, and EWC and our method require additional memory as much as the size of the network.
    The proposed method utilizes the information stored in the network and achieve best performance with a small fixed memory.
    }
    \label{fig:results_memory}
    \vspace{-1.4em}
\end{figure}

\begin{figure*}[t]
    \vspace{-0.7em}
    \includegraphics[width=1.0\textwidth]{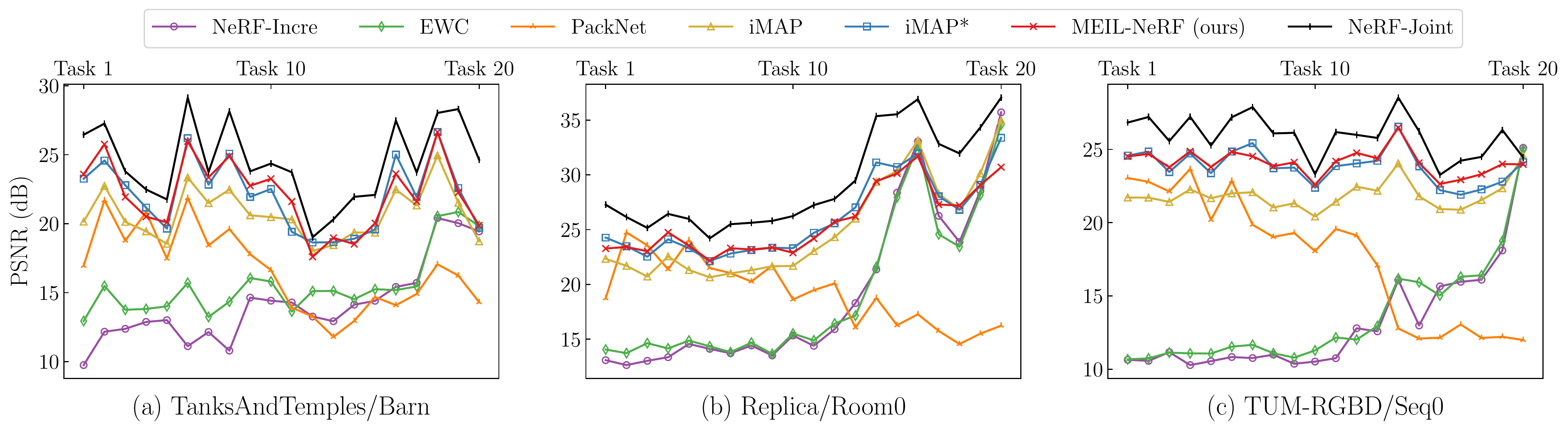}
    \vspace{-1.8em}
    \caption{\textbf{Evaluation of All Tasks.} 
    The figure shows the evaluation of all tasks after training the last task in terms of PSNR.
    The baseline (NeRF-Incre) shows significant performance degradation in the initial tasks. 
    EWC shows no improvement, and PackNet shows performance degradation as the number of available parameters decreases.
    For iMAP* (similar performance to ours), the memory increases quickly.
    iMAP (similar memory to ours) shows low performance. 
    Note that we randomly select one from each dataset and display MS-SSIM results in supple.
    }
    \label{fig:results_quan3}
    \vspace{-1.em}
\end{figure*}

\begin{figure*}[t]
    \includegraphics[width=1.0\textwidth]{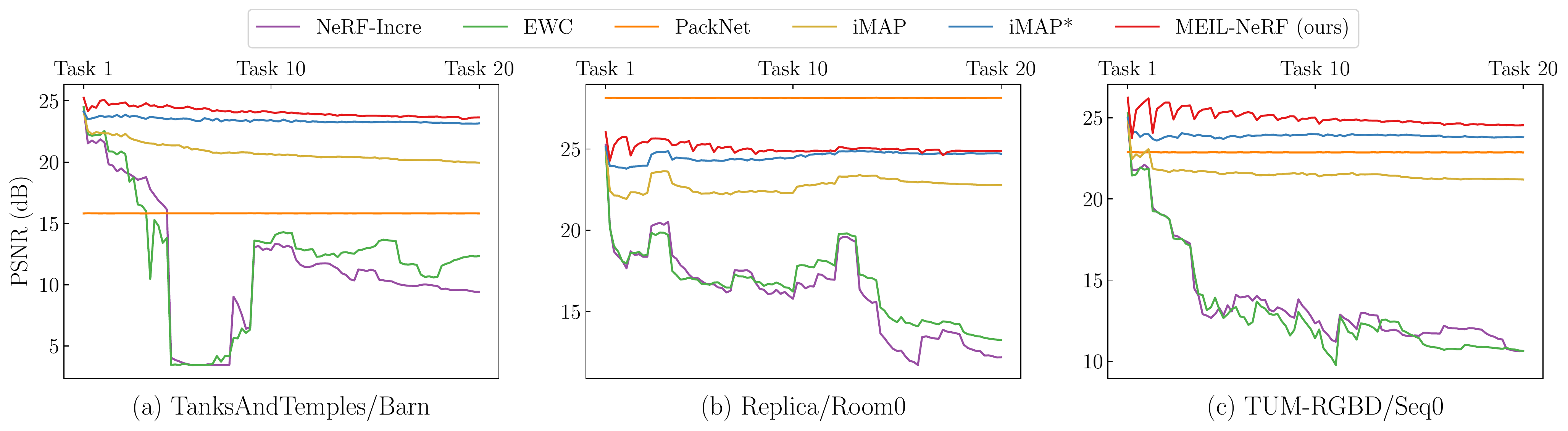}
    \vspace{-1.8em}
    \caption{\textbf{Performance Degradation of the First Task.}
    The figure displays the performance degradation of the first task (Task 1) as the network continuously learns all the tasks.
    In the baseline (NeRF-Incre), the network rapidly forgets the first task.
    EWC is simliar to the baseline, and PackNet shows no forgetting because the parameters of the first task are fixed.
    iMAP exhibit a stable performance after a slight drop in early stage, due to the lack of exemplar.
    In contrast, iMAP* and our method significantly decelerate forgetting speed, and even our method slightly outperforms iMAP*.
    The fluctuation in our method is due to the $\lambda_p$ scheduling which is described in supple.
    We display MS-SSIM results in supple due to lack of space.
    }
    \label{fig:fig_inittask}
    \vspace{-1.4em}
\end{figure*}

\subsection{Experiment Settings} \label{section:exp_expsettings}
We set the vanilla NeRF under incremental scenario as a baseline and name it as \textit{NeRF-Incre}.
NeRF-Incre is incrementally trained with only current task data, making it susceptible to catastrophic forgetting and serving as lower bound.
We also train the vanilla NeRF with the standard joint training, where all data is made available at all times.
We label it as \textit{NeRF-Joint} and regard it as upper bound.

We also compare against the representative incremental learning algorithms (regularization, parameter isolation, and replay) on NeRF to observe how the algorithms developed for classification tasks generalize to NeRF under incremental scenarios.
Selected as the representative of regularization, \textit{Elastic Weight Consolidation} (EWC)~\cite{kirkpatrick2017overcoming} penalizes the changes in parameters that are important for past tasks.
For parameter isolation, we adopt the idea of \textit{PackNet}~\cite{mallya2018packnet}, which prunes less important parameters and re-train to make room for the next task. 
As for replay, we follow the general implementations of \textit{iMAP}~\cite{sucar2021imap}, which has tailored replay algorithm to SLAM.
At the end of each task, we store random exemplars weighted by its loss, and repeat them while learning a new task.
Since the performance of replay-based algorithms depends on the amount of memory reserved for each task, we vary the amount of memory size and implement two versions: iMAP that has  a similar memory usage to our method and iMAP* that has similar performance to ours without restricting the memory usage.

We set the number of images $\mathcal{N} = 5$ for each task by default and run $3000*\mathcal{N}$ iterations per task. 
We use $m_c = 4096$ rays for learning the current task at every iteration. 
Our method and iMAP prevent forgetting by also training on $m_p = m_c/2$ past rays, which are either generated (our method) or sampled from the memory buffer (iMAP).

\begin{figure*}[t]
    \includegraphics[width=\textwidth]{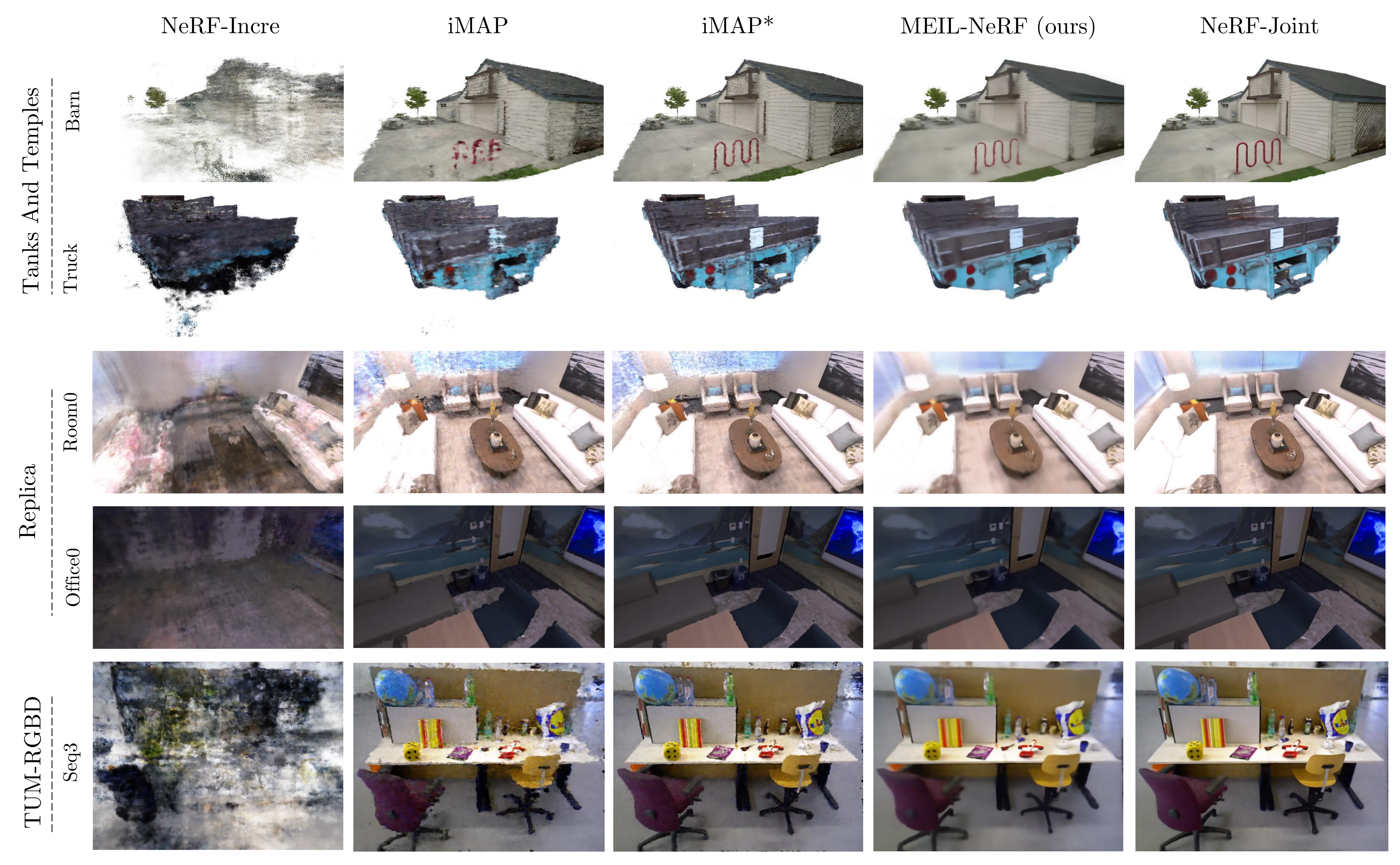}
    \vspace{-1.8em}
    \caption{\textbf{Qualitative Results.} 
    We show the reconstructed images for the early task after training of all tasks in an incremental manner.
    The baseline (NeRF-Incre) severely suffers catastrophic forgetting and forgets the early-task scene information, whereas iMAP, iMAP*, and our method preserves the scene of interest.
    iMAP and iMAP* are our reproduction such that iMAP has similar memory usage to ours, while iMAP* is produced to have similar performance to ours but without memory constraint (please refer to Figure~\ref{fig:results_memory}).
    iMAP shows many artifacts due to the lack of samples.
    iMAP* with more memory displays improved sharp appearance, but still has distortion artifacts. 
    Despite using the similar memory usage to iMAP, our MEIL-NeRF shows more plausible and visually pleasing results than iMAP.
    Furthermore, the reconstruction quality of MEIL-NeRF is similar to iMAP*, despite using substantially less memory than iMAP*.
    Slight blur in the reconstruction by MEIL-NeRF may be due to using diverse generated samples (but with the same number of samples as iMAP*) we obtain from RGN and NeRF.
    Please refer to the supplementary material to see more qualitative results on EWC and PackNet and other data sequences. 
    }
    \label{fig:results_qual}
    \vspace{-1.4em}
\end{figure*}

\subsection{Datasets} \label{section:exp_dataset}
We construct datasets for incremental scenarios, based on \textit{Tanks and Temples}~\cite{Knapitsch2017}, \textit{Replica}~\cite{straub2019replica}, and \textit{TUM-RGBD}~\cite{sturm2012benchmark}. 
To simulate the incremental scenarios, we rearrange the order of camera such that it moves sequentially and we select a portion of the dataset such that the previous images are not revisited.
Tanks and Temples consists of photos taken while moving around a scene or object of interest. 
The dataset consists of scenes of various sizes ranging from a large scene (e.g., Barn) to a small object (e.g., Family).
We rearrange the camera to turn in one direction, ending the sequence before it reaches the first scene.
Replica dataset is a synthetic indoor dataset with accurate camera poses and rendered images. 
Since the camera moves smoothly, we sub-sample the images so that there is enough change in degrees between tasks.
As with Tanks and Temples, we cut a part of the camera sequence, so the network does not review the past images.
TUM-RGBD dataset is a dataset made for SLAM, which was taken with a hand-held camera or a camera mounted on a robot. 
Since images are taken while moving, most sequences have severely blurry pictures.
For better evaluation of catastrophic forgetting, we select sequences with a low degree of blur.

\subsection{Results} \label{section:exp_results}
We evaluate the models with respect to Peak Signal-to-Noise Ratio (PSNR) and MultiScale Structural Similarity Index Measure (MS-SSIM). 
We show the average performance of all tasks on Table~\ref{tab:dataset_quan} and the performances of each task on three randomly selected scenes in Figure~\ref{fig:results_quan3}.
Figure~\ref{fig:fig_inittask} also displays how the performance of the first task (Task 1) changes as the models learn new tasks over time to show catastrophic forgetting.
As expected from NeRF-Incre, the performance on the initial task severely deteriorates due to catastrophic forgetting.
EWC fails to reduce the adverse effects of catastrophic forgetting.
In fact, EWC is shown to even hinder the learning of new tasks, decreasing the average performance over all tasks.
PackNet performs worse on newer tasks since the number of parameters assigned to a new task reduces gradually.
On the other hand, PackNet maintains the performance on the initial task since the parameters for the first task are fixed after pruning.
Despite directly saving past tasks data, iMAP performs lower than our MEIL-NeRF, which uses similar amount of memory.
With constant memory usage, our MEIL-NeRF effectively mitigates catastrophic forgetting and performs on par with iMAP*, whose memory usage is substantially larger than our method, as shown in Figure~\ref{fig:results_memory}. 
This demonstrates the efficiency and effectiveness of using NeRF as a memory storage and retrieving past information from NeRF itself.

We can gain more insight from observing the qualitative results shown in Figure~\ref{fig:results_qual}.
The figure shows the reconstruction of images for early tasks after incremental learning has finished (i.e., all tasks have been learned sequentially).
The replay-based method iMAP* shows relatively sharp images, but there are some distortion artifacts.
Since only selected samples are stored and repeatedly used, iMAP* overfits to specific rays, compromising multiview consistency.
On the contrary, MEIL-NeRF shows smoother reconstruction without the artifacts.
Our method has less artifacts since it can generate various samples by using RGN and NeRF.
The reconstruction by MEIL-NeRF may have gotten a smoothing effect affected by accumulated errors from the iterative distillation process.
Selecting important generated samples for training NeRF to obtain sharper and more accurate reconstruction can be one of the interesting future research directions.

\subsection{Ablations} \label{section:exp_ablation}
Table~\ref{tab:ablation_methodmc} shows how performance changes with the current task batch size $m_c$.
Compared to iMAP*, our method deteriorates relatively quickly as the batch size becomes smaller.
The smaller batch size leads to underfitting of each task.
Thus, the retrieved past task information from the network can become more inaccurate, degrading the performance.

\begin{figure}[t]
    \includegraphics[width=1.0\columnwidth]{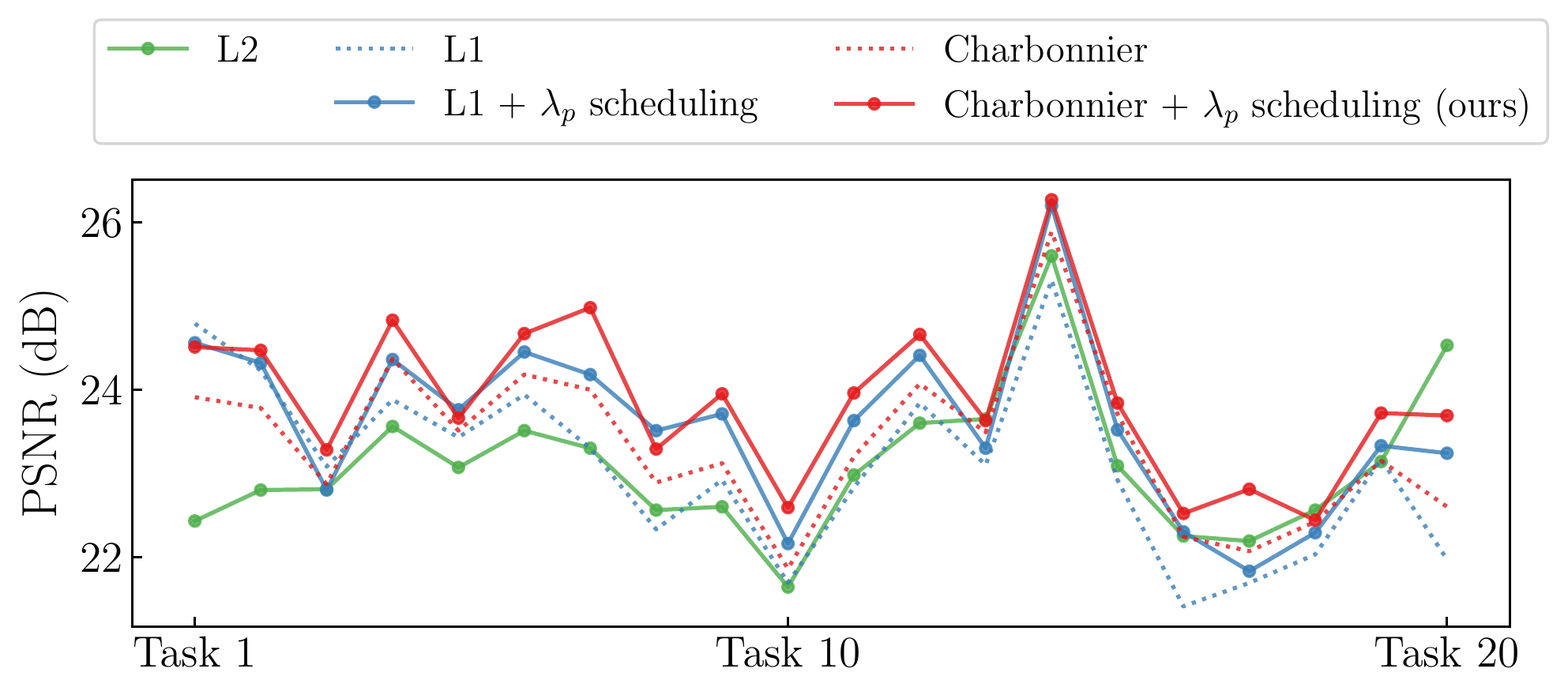}
    \caption{\textbf{Ablation for Loss.} 
    We show TUM-RGBD/Seq0 results with L2, L1 and Charbonnier loss.
    The edge-preserving property of L1 and Charbonnier loss improve the performance especially in MS-SSIM.
    $\lambda_p$ scheduling also helps improve performance by organizing a learning curriculum, from learning new tasks to remembering past information.
    }
    \label{fig:ablation_Loss}
    \vspace{-0.6em}
\end{figure}

We also perform an ablation study on loss function used for past tasks to justify our choice of Charbonnier loss function, as shown in Figure~\ref{fig:ablation_Loss}.
L1 and Charbonnier loss exhibit higher performances than L2 loss, while $\lambda_p$ scheduling provides further improvement.
L1 loss and its variants (e.g., Charbonnier) known to facilitate the learning of networks to preserve edges and sharpness in reconstructed images, demonstrating even more performance improvement in MS-SSIM, as shown in the supplementary material. 

\input{sections/table_ablation_methodmc}

\begin{figure}[t]
    \includegraphics[width=1.0\columnwidth]{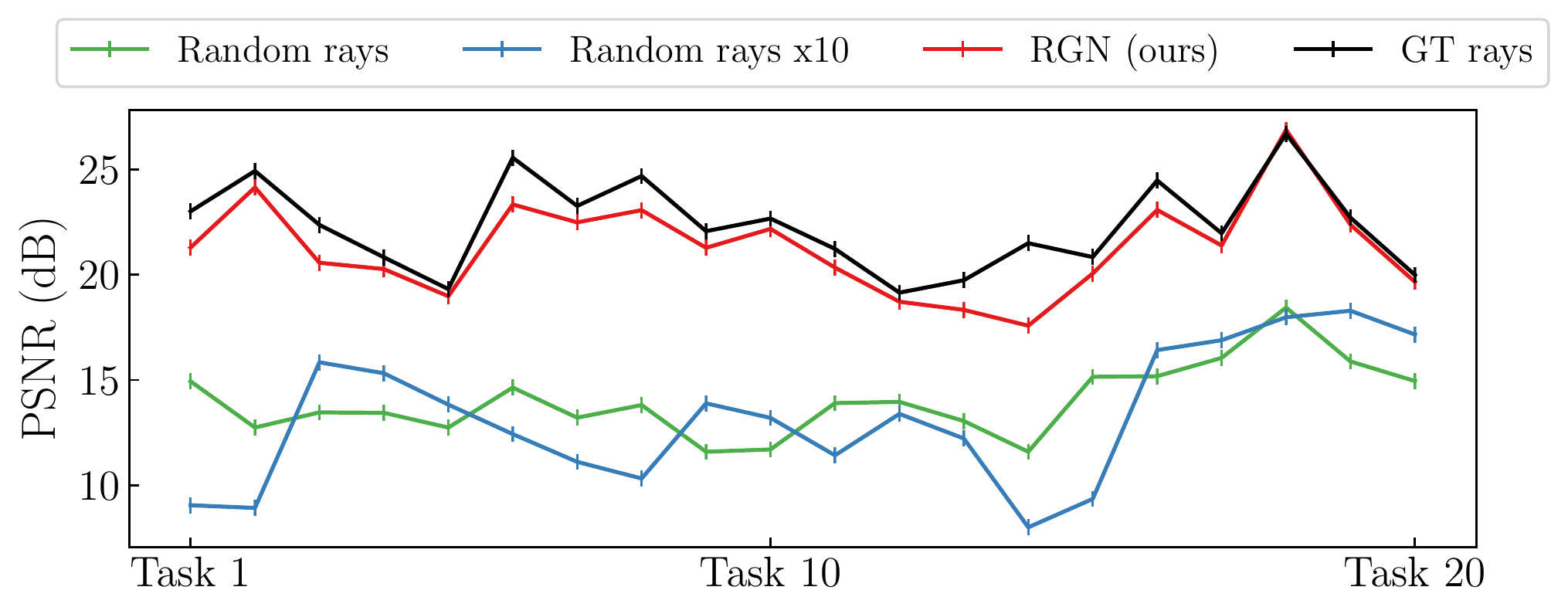}
    \vspace{-1.8em}
    \caption{\textbf{Ablation for Ray Generator Network.} 
    We show Tanks and Temples/Barn results with GT past rays, rays from RGN, and rays cast randomly.
    Exhibiting remarkably similar performance to using GT past rays, RGN demonstrates the capability of remembering past rays without increasing memory.
    Random rays still perform poorly even with using ten times more rays.
    }
    \label{fig:ablation_RGN}
    \vspace{-1.1em}
\end{figure}

Figure~\ref{fig:ablation_RGN} reports the ablation experiments on ray generator network (RGN).
The result shows that past color information cannot be retrieved from NeRF if random rays are used (even with 10 times more rays), underlining the importance of using desirable rays that point at the scene.
To show the effectiveness of RGN in remembering past rays, we compare against using GT past rays.
RGN remarkably provides similar performance to using GT past rays, underscoring the effectiveness of RGN in remembering past rays with constant memory usage.

%% file: sections/table_ablation_methodmc.tex
\begin{table}[t]
\centering
\caption{Ablation for batch size of current task on TUM-RGBD/Seq0 ($m_p=m_c/2$)}
\label{tab:ablation_methodmc}
\resizebox{\linewidth}{!}{
\large{
\begin{tabular}{|c|cccc|c|}
\hline
\multicolumn{1}{|c|}{\multirow{1}{*}{\small{PSNR / MS-SSIM}}} & $m_c=512$ & $m_c=1024$ & $m_c=2048$ & $m_c=4096 $ & $\Delta$ \\ \hline
\multicolumn{1}{|c|}{iMAP*} &
  \textbf{23.69} / \textbf{0.867} &
  \textbf{23.78} / \textbf{0.871} &
  \textbf{23.98} / \textbf{0.880} &
  23.93 / 0.878 &
  0.24 / 0.011 \\
\multicolumn{1}{|c|}{MEIL-NeRF} &
  22.89 / 0.850 &
  23.44 / 0.866 &
  23.77 / 0.874 &
  \textbf{24.12} / \textbf{0.882} &
  1.23 / 0.032 \\ \hline
\multicolumn{1}{|c|}{NeRF-Joint} &
  25.31 / 0.909 &
  25.63 / 0.914 &
  25.93 / 0.919 &
  26.10 / 0.922 & 
  0.79 / 0.013 \\ \hline
\end{tabular}}
}
\vspace{-1.6em}
\end{table}

%% file: sections/6limitation7conclusion.tex
\section{Limitation}
Since MEIL-NeRF uses NeRF as a memory storage and query NeRF for past information, it has similar high-latency limitations as NeRF.
NeRF has high-latency due to the procedure of sampling points and integrating them along camera rays.
We believe that latency problems can be overcome through employing recent works~\cite{muller2021real, mueller2022instant} that have attempted to improve the latency of NeRF.

\section{Conclusion}
In this work, we aim to further push NeRF towards practical scenarios, where we focus on incremental scenario.
Under the formulated incremental scenario, previous incremental learning algorithms have failed to work for NeRF or demand for a large amount of memory.
Thus, we propose a \textbf{M}emory-\textbf{E}fficient \textbf{I}ncremental \textbf{L}earning algorithm for \textbf{NeRF} (\textbf{MEIL-NeRF}).
MEIL-NeRF manages to alleviate catastrophic forgetting with constant memory usage by treating the network itself as a mean of storing past information.
We hope this work can inspire future research works on NeRF in incremental learning scenarios.

%% file: sections/8suppleWithoutFakeref.tex
In this supplementary document, we show the results and ablations omitted from the main paper due to lack of space and describe the details of comparison methods.
Section~\fakeref{S1} reports quantitative comparison results in terms of MS-SSIM; Section~\fakeref{S2} exhibits qualitative results on EWC and PackNet and other data sequences;
Section~\fakeref{S3} includes the additional ablation studies for the proposed method;
Section~\fakeref{S4} elaborates on the details of the ray generator network (RGN);
Finally, Section~\fakeref{S5} provides implementation details and ablation studies on hyperparameters for other methods.

\section{Quantitative Results} \label{section:supp_quan}
\begin{figure*}[t!]
    \centering
    \includegraphics[width=\linewidth]{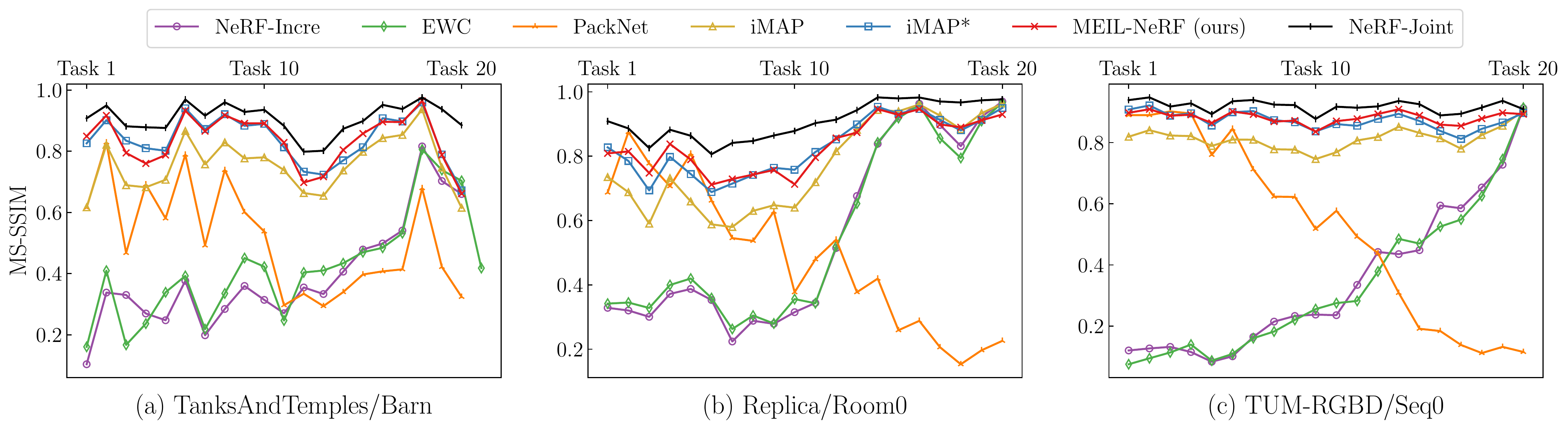}
    \caption{\textbf{Evaluation of All Tasks.} 
    The figure shows the MS-SSIM evaluation averaged over all tasks after the incremental learning of all tasks, providing the MS-SSIM counterpart of Figure~\ref{fig:results_quan3}.
    } 
    \label{fig:supp_quan_msssim}
\end{figure*}

\begin{figure*}[t!]
    \centering
    \includegraphics[width=\linewidth]{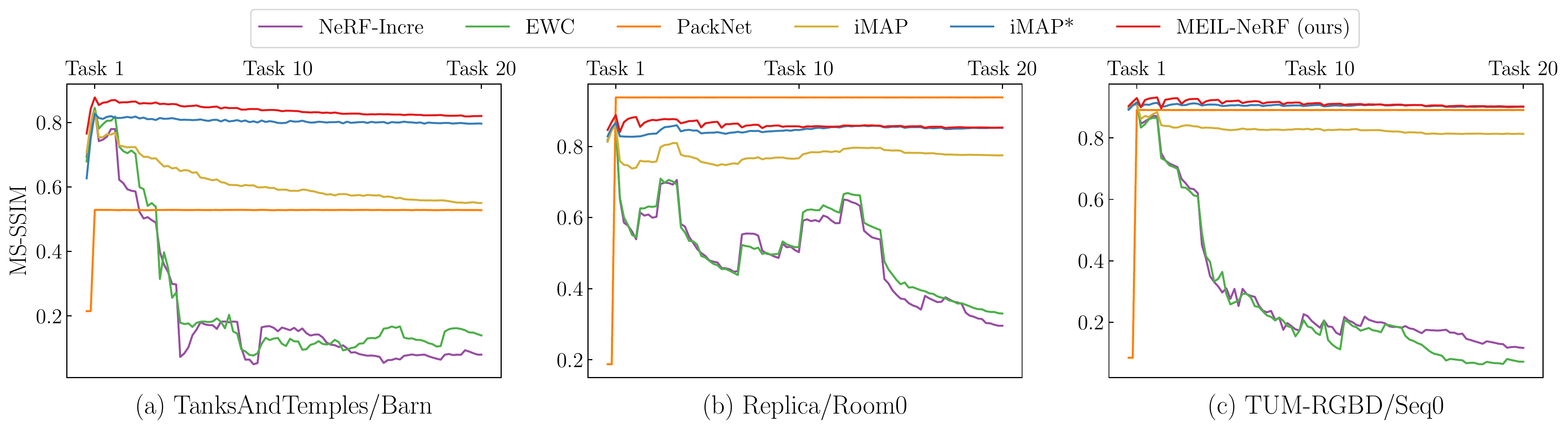}
    \caption{\textbf{Performance Degradation of the First Task.}
    The figure shows the MS-SSIM performance degradation (and thus catastrophic forgetting) of the first tasks (Task 1) after the incremental learning of all tasks, providing the MS-SSIM counterpart of Figure~\ref{fig:fig_inittask}.
   }
   \label{fig:supp_inittask_msssim}
\end{figure*}

We show the MS-SSIM evaluation of all tasks after training is finished in Figure~\ref{fig:supp_quan_msssim} and the performance degradation of the first task (Task 1) in Figure~\ref{fig:supp_inittask_msssim}, in terms of MS-SSIM.
They show similar trends to the PSNR evaluation in Figure~\ref{fig:results_quan3} and Figure~\ref{fig:fig_inittask} of the main paper, respectively.

\begin{figure}[t]
    \centering
    \includegraphics[width=\linewidth]{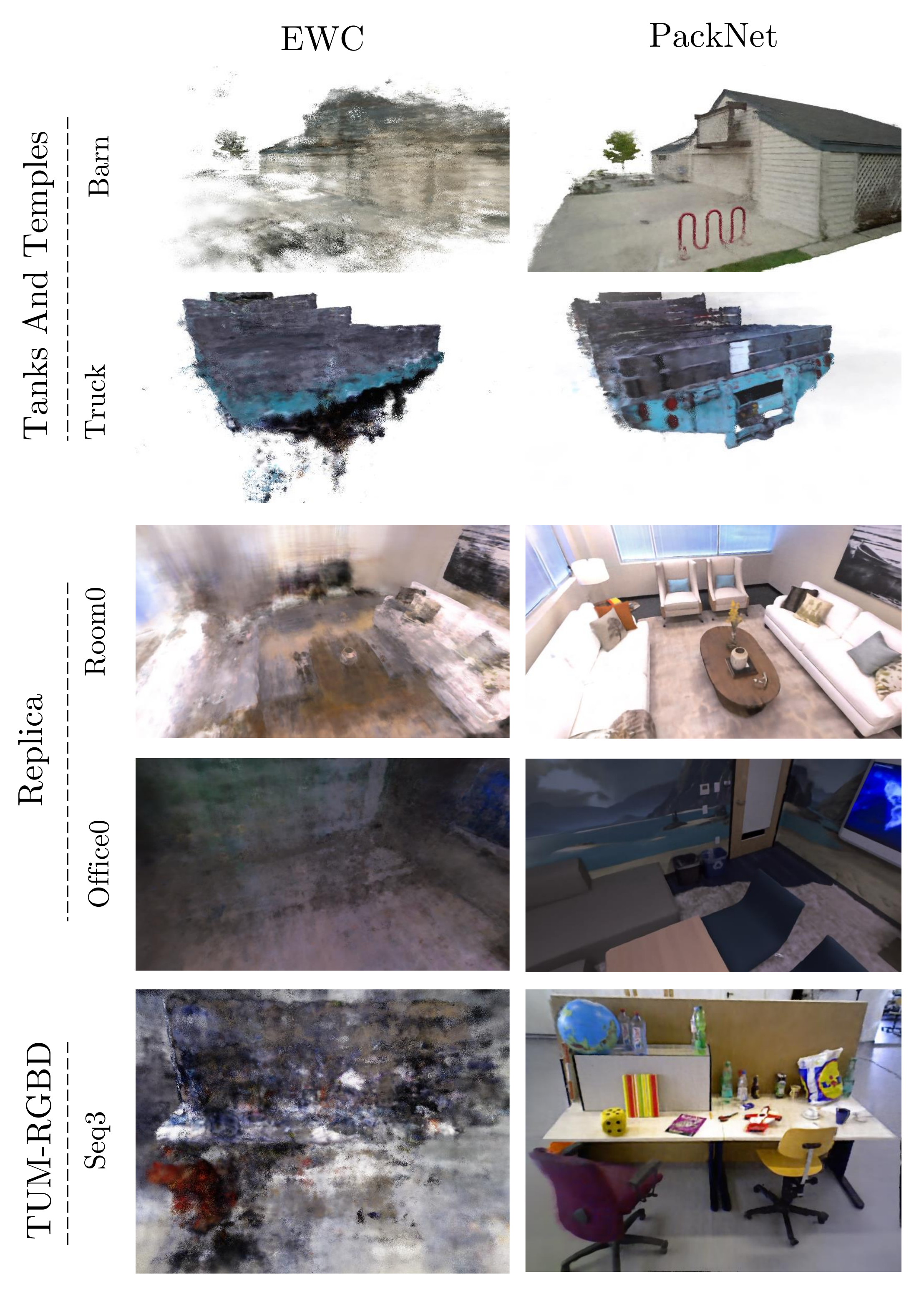}
    \vspace{-2.0em}
    \caption{\textbf{Qualitative Results.} 
    We visualize the images reconstructed by EWC and PackNet for the early task after the incremental learning of all tasks, omitted in Figure~\ref{fig:results_qual}.
    While PackNet provides high quality in earlier tasks, it gives poor quality in newer tasks since very few parameters are reserved for learning new tasks, as shown in Figure~\ref{fig:supp_data_sequence_barn}~-~\ref{fig:supp_data_sequence_seq0}.}
    \label{fig:supp_additional_qual}
    \vspace{-1.5em}
\end{figure}

\section{Qualitative Results} \label{section:supp_qual}
\begin{figure*}
    \centering
    \includegraphics[width=\linewidth]{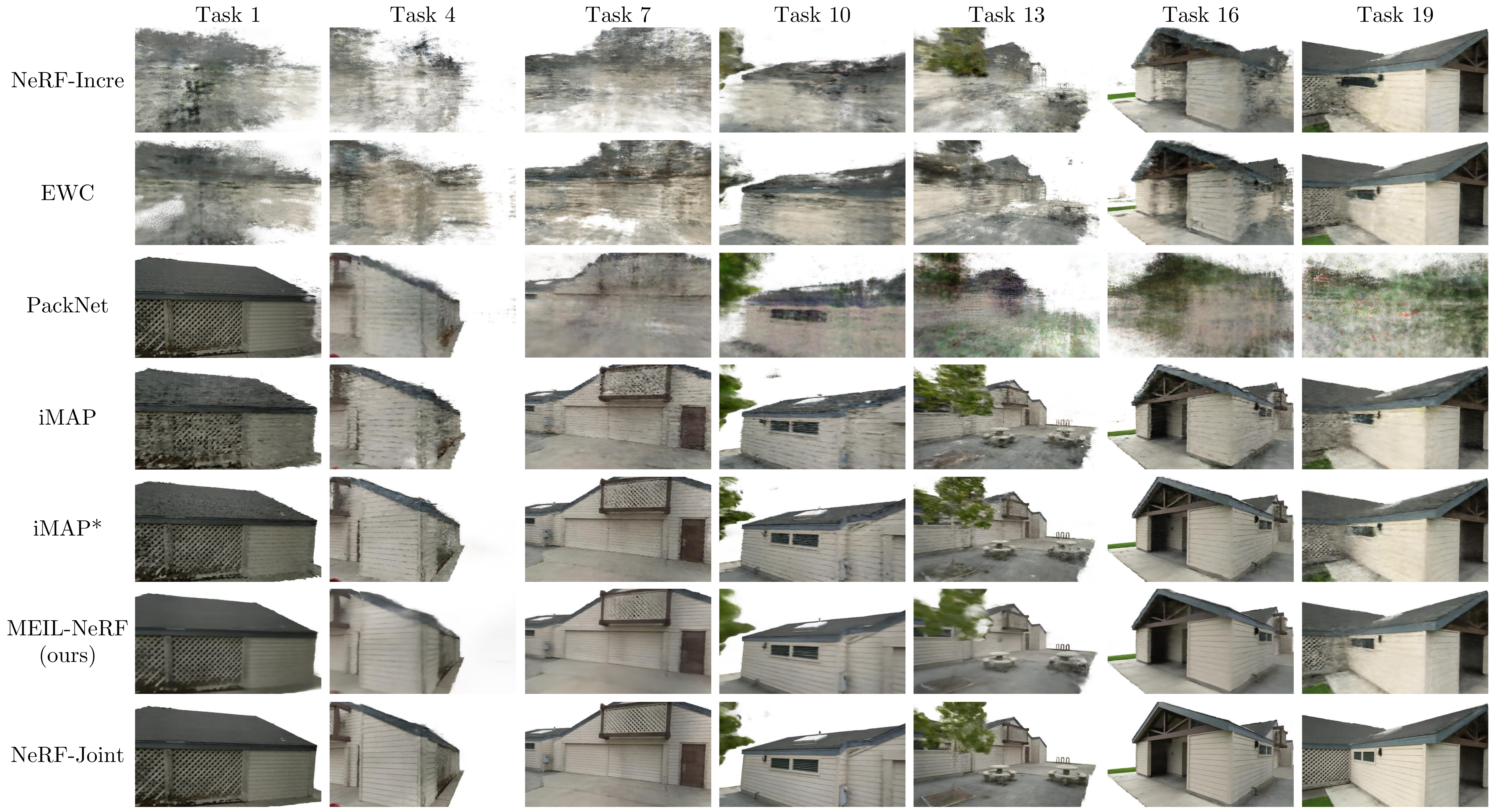}
    \vspace{-1.8em}
    \caption{\textbf{Qualitative Results in Tanks And Temples/Barn.}}
    \label{fig:supp_data_sequence_barn}
    \vspace{-1.8em}
\end{figure*}

\begin{figure*}
    \centering
    \includegraphics[width=\linewidth]{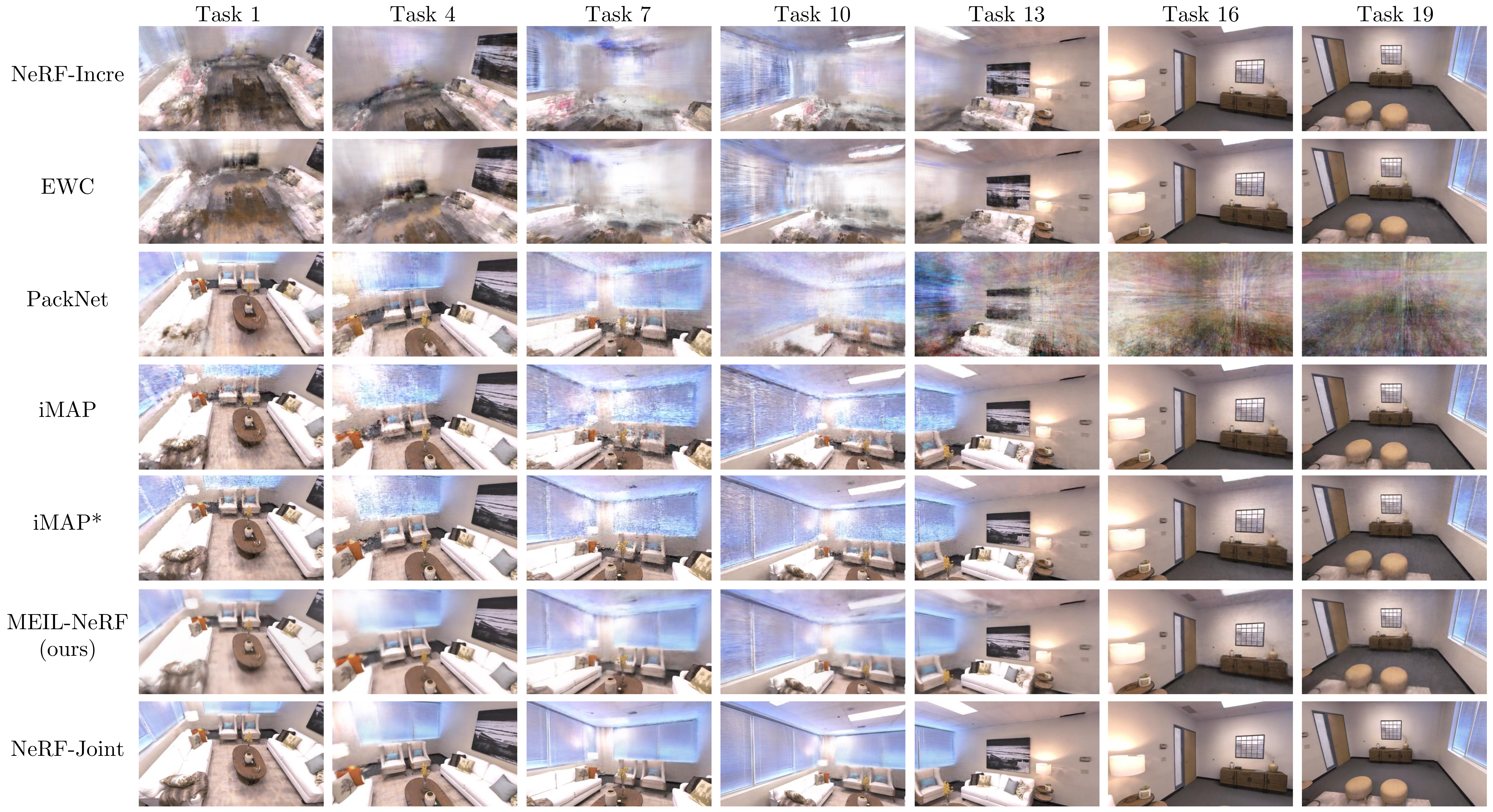}
    \vspace{-1.8em}
    \caption{\textbf{Qualitative Results in Replica/Room0.}
    }
    \label{fig:supp_data_sequence_room0}
    \vspace{-1.8em}
\end{figure*}

\begin{figure*}[t!]
    \centering
    \includegraphics[width=\linewidth]{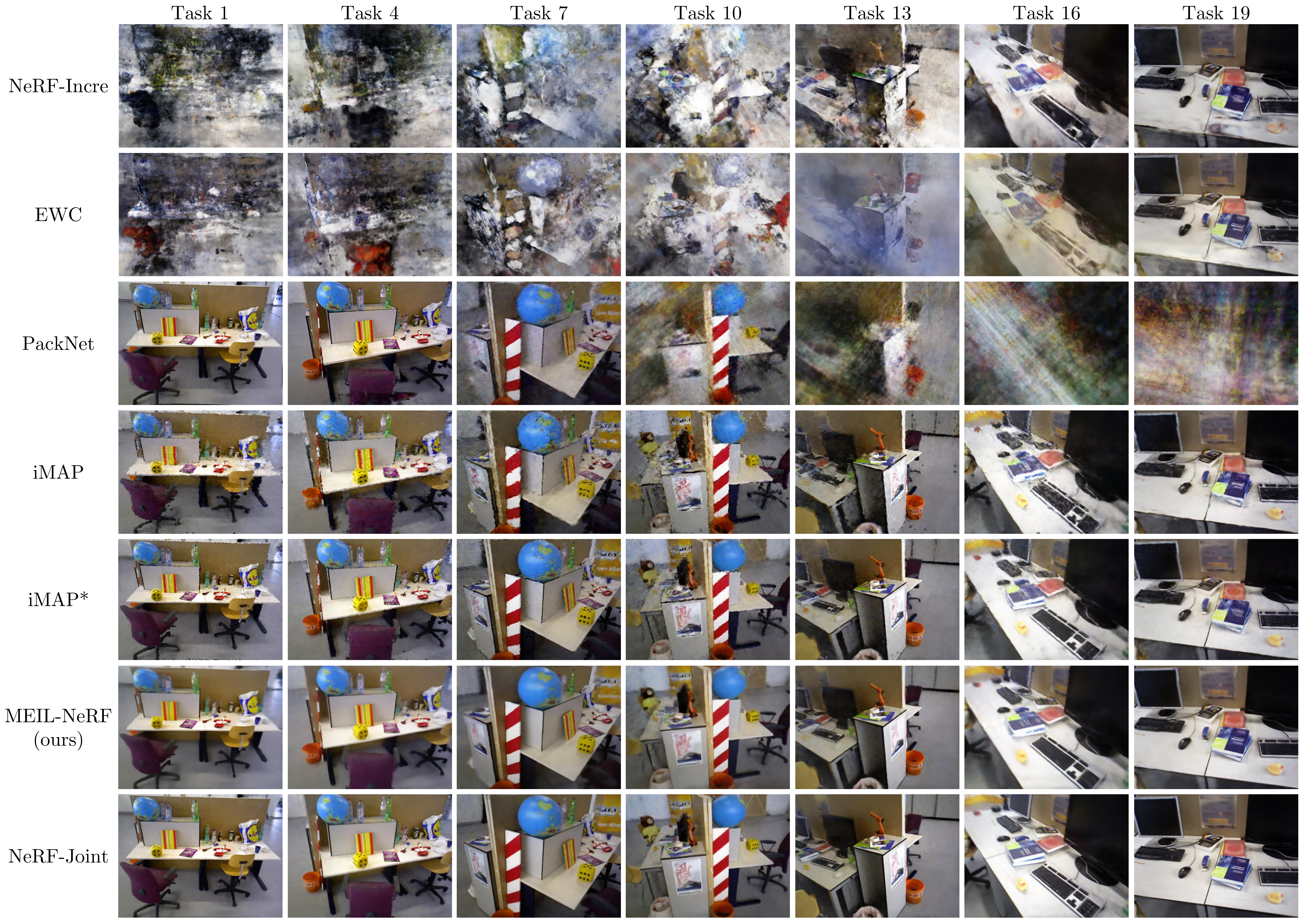}
    \vspace{-1.8em}
    \caption{\textbf{Qualitative Results in TUM-RGBD/Seq0.}}
    \label{fig:supp_data_sequence_seq0}
    \vspace{-1.8em}
\end{figure*}

Figure~\ref{fig:supp_additional_qual} shows the qualitative results of EWC and PackNet omitted from Figure~\ref{fig:results_qual}.
EWC exhibits severe catastrophic forgetting, similar to NeRF-Incre.
PackNet has high reconstruction quality in the early task like NeRF-Joint, since the parameters for the learned tasks are fixed after pruning and re-training.
However, PackNet fails to provide high reconstruction quality in newer tasks, as discussed with additional qualitative results below.
We also present other qualitative results in Figure~\ref{fig:supp_data_sequence_barn}, Figure~\ref{fig:supp_data_sequence_room0}, and Figure~\ref{fig:supp_data_sequence_seq0} displaying the images in some tasks after training is finished.
In essence, the figures are the qualitative counterparts of Figure~\ref{fig:results_quan3} and Figure~\ref{fig:supp_quan_msssim}.
NeRF-Incre and EWC demonstrate severe catastrophic forgetting.
In contrast, PackNet performs worse on newer tasks since the number of available parameters for learning new tasks gradually decreases.
iMAP and iMAP* prevent catastrophic forgetting to a great extent and provide good reconstruction quality across tasks; however, many blurs and distortions are present in the reconstructed images.
Our method shows relatively smoothed images without noticeable artifacts.

\begin{figure}
    \centering
    \includegraphics[width=\linewidth]{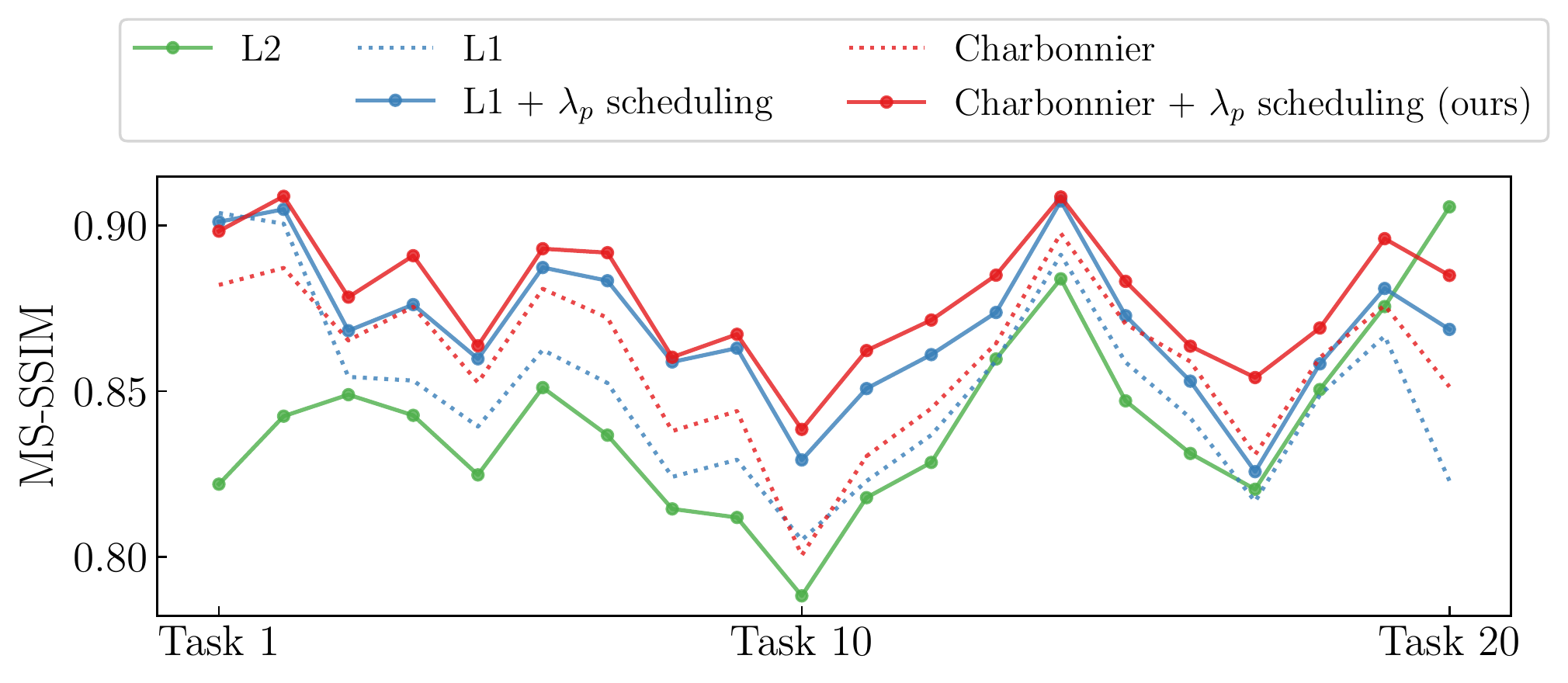}
    \vspace{-2.0em}
    \caption{\textbf{Ablation for Loss.} 
    This figure provides the MS-SSIM counterpart of Figure~\ref{fig:ablation_Loss}.
    We show the MS-SSIM evaluation on TUM-RGBD/Seq0 sequence after training our model with L2, L1, and Charbonnier loss to preserve the knowledge of past tasks.
    L1 loss and its variants (e.g., Charbonnier) demonstrate even more performance improvement in MS-SSIM.
    The result corroborates the previous finding that L1 loss and its variants lead to better edge-preserving capability.}
    \label{fig:supp_ablation_loss_msssim}
    \vspace{-0.5em}
\end{figure}
\input{sections/supp_table_ablation_mcmp.tex}
\begin{figure}[t]
    \centering
    \includegraphics[width=\linewidth]{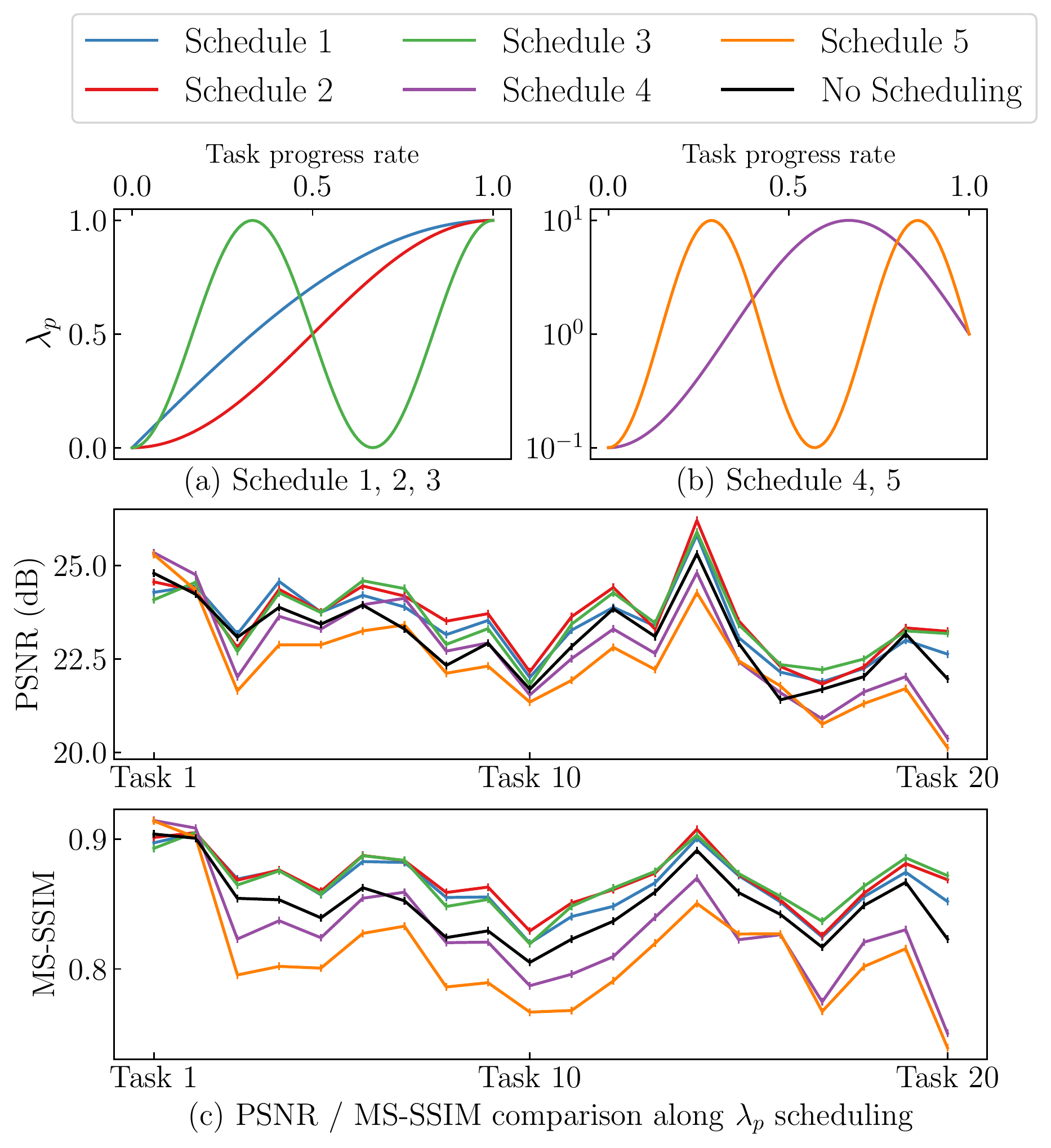}
    \vspace{-1.0em}
    \caption{\textbf{Hyperparameter Scheduling.} 
    We show the scheduling examples for the hyperparameter $\lambda_p$ in (a) and (b).
    We present the performance of each schedule in (c), evaluated in TUM-RGBD/Seq0.
    We use schedule 2, showing slightly better results among them.}
    \label{fig:supp_ablation_lambdap_scheduling}
    \vspace{-1.0em}
\end{figure}

\section{Ablations for the proposed method} \label{section:supp_ablation}
\noindent
\textbf{Loss function used for past tasks.}
In Figure~\ref{fig:supp_ablation_loss_msssim}, we conduct an ablation study on the loss function used for remembering past tasks in terms of MS-SSIM. 
L1 loss and its variants provide more notable performance improvement in MS-SSIM, compared to the PSNR results shown in Figure~\ref{fig:ablation_Loss}.
Considering that MS-SSIM measures the structural similarity, the result corroborates the previous finding that L1 and Charbonnier loss provide better edge-preserving capability, reducing the errors from the iterative process of self-distillation.

\noindent
\textbf{Hyperparameter $\lambda_p$ scheduling.}
A hyperparameter $\lambda_p$ in Equation~\eqref{eq:method_loss} controls a trade-off between learning a new task and remembering past tasks.
So we establish a learning curriculum for the hyperparameter during the learning of each task, which guides NeRF to focus on learning a new task first and then gradually shift the focus to remembering past tasks.
We implement the learning curriculum by scheduling the hyperparameter $\lambda_p$ according to the progress rate of learning in one task.
Figure~\ref{fig:supp_ablation_lambdap_scheduling} shows the PSNR and MS-SSIM assessment of several designs of $\lambda_p$ scheduling.
For the task progress rate $r$, the schedule $1\sim5$ are implemented as,
\begin{align}
    \lambda_p&=\cos{\left(\frac{\pi}{2}(1-r)\right)}, \\ 
    \lambda_p&=\frac{1}{2}\cos{\left(\pi(1+r)\right)}, \\ 
    \lambda_p&=\frac{1}{2}\cos{\left(\pi(1+3r)\right)}, \\ 
    \lambda_p&=10^{\cos{\left(\pi(1+\frac{3}{2}r)\right)}}, \\ 
    \lambda_p&=10^{\cos{\left(\pi(1+\frac{7}{2}r)\right)}},
\end{align}
respectively.
Schedule $1\sim3$, which progressively climb from zero to one, generally perform better than Schedule $4\sim5$.
We select the second one (schedule 2), which demonstrates slightly better results among Schedule $1\sim3$.

\begin{figure}[t!]
    \centering
    \includegraphics[width=\linewidth]{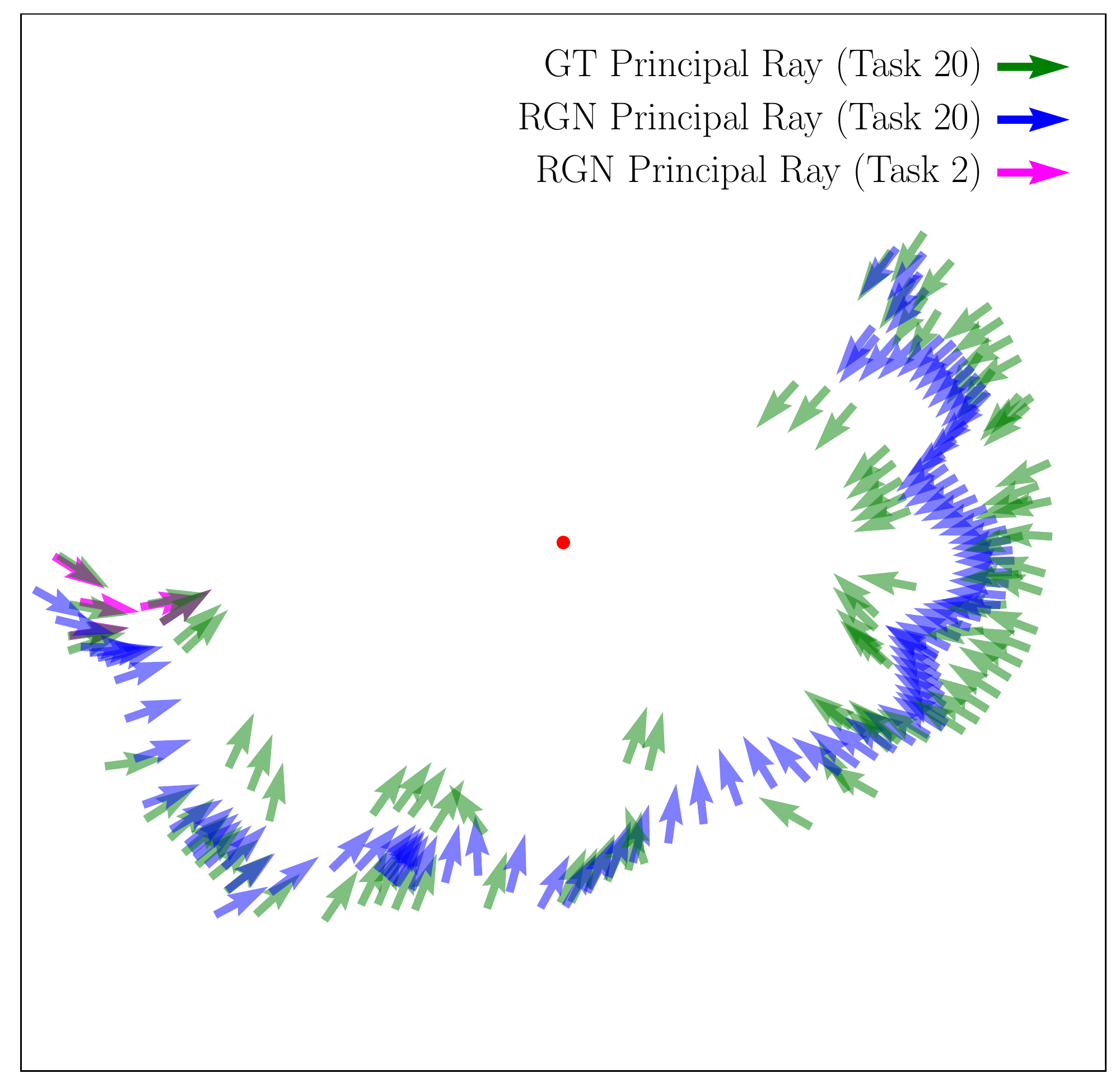}
    \vspace{-1.0em}
    \caption{\textbf{Visualization of GT rays and RGN generated rays.} 
    We show the GT principal rays and RGN-generated principal rays in the Tanks and Temples/Barn dataset.
    The tail and direction of the arrow indicate the camera origin and direction, respectively.
    For convenience, we present the location of the scene with a red dot.
    }
    \label{fig:supp_fig_ablation_RGN}
    \vspace{-1.0em}
\end{figure}
\begin{figure}[t]
    \centering
    \includegraphics[width=\linewidth]{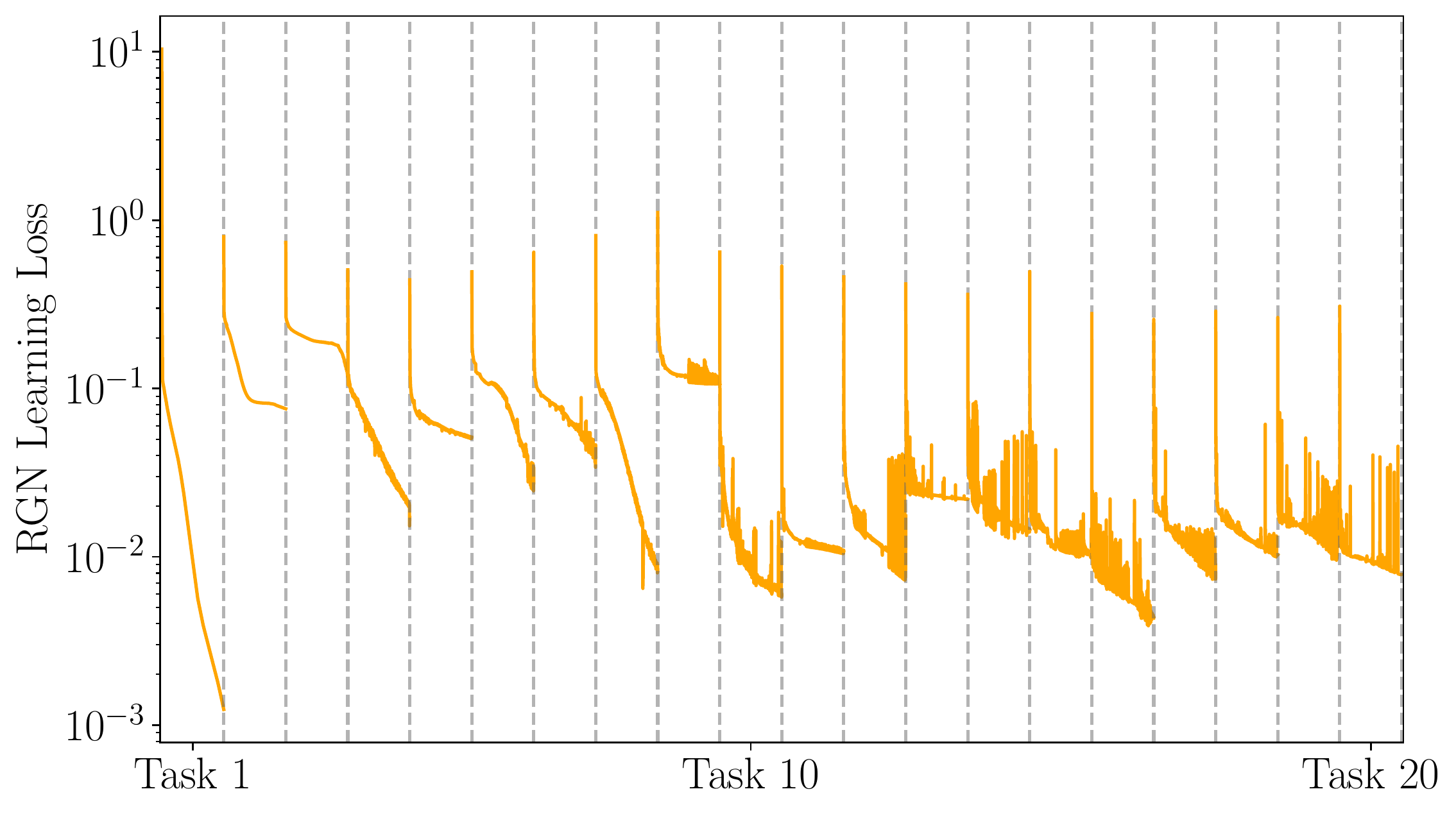}
    \vspace{-1.0em}
    \caption{\textbf{RGN Training Loss.} 
    We show how the average RGN training loss for each task changes as the number of tasks increases.
    Although RGN is a tiny network, it preserves the past ray information well by reaching a certain level of loss.
    }
    \label{fig:supp_fig_ablation_RGNloss}
    \vspace{-1.0em}
\end{figure}

\noindent
\textbf{Batch size of past rays $m_p$.}
Table~\ref{tab:supp_ablation_mcmp} shows the change in the overall performance according to the batch size for past task $m_p$.
In general, the performance tends to increase with $m_p$.
Considering the trade-off between the performance and the computational efficiency, we choose $m_p=m_c/2$.

\begin{figure*}[t!]
    \centering
    \includegraphics[width=\linewidth]{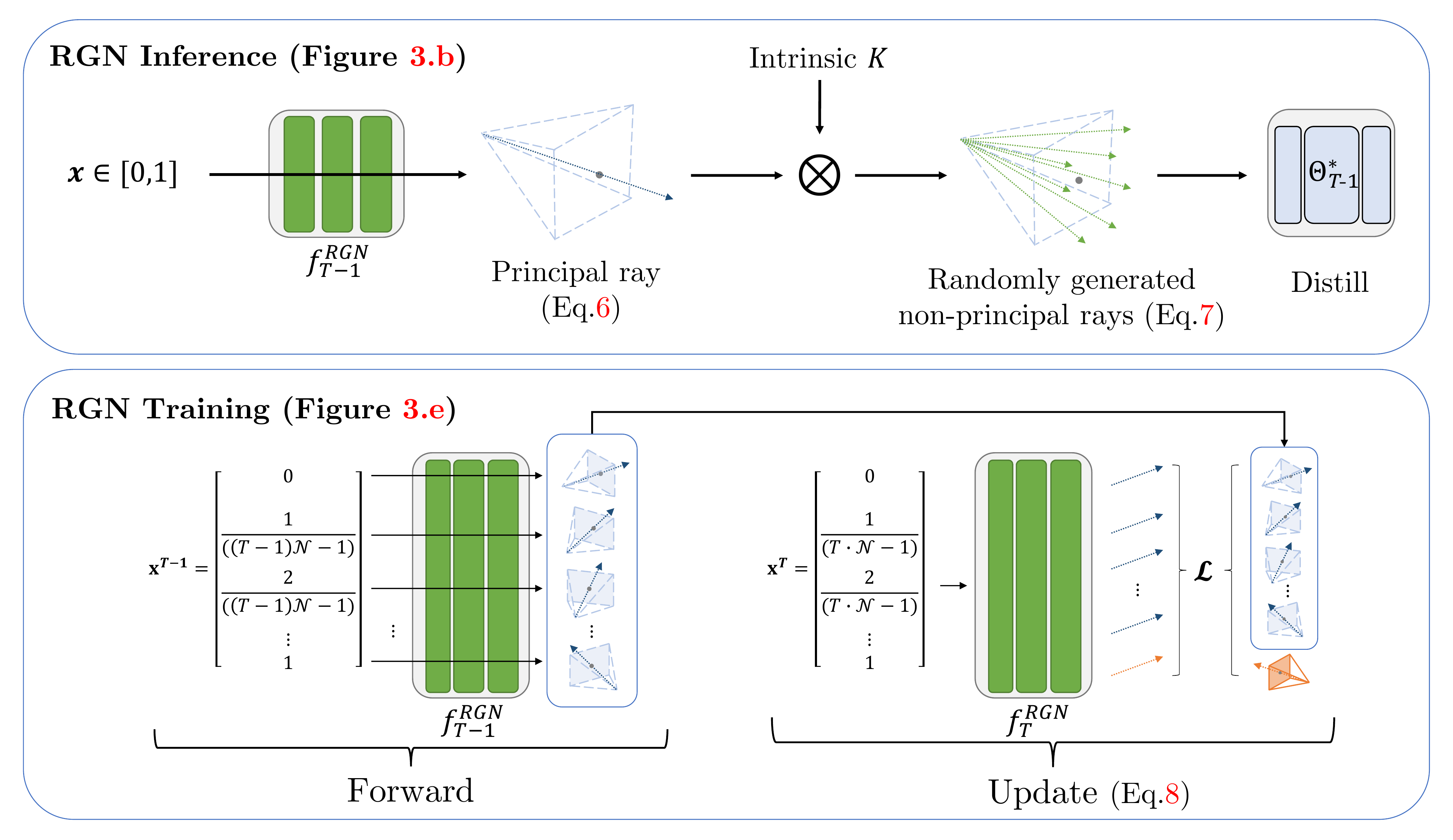}
    \vspace{-1.0em}
    \caption{\textbf{Details of ray generator network.} 
    We visualize the details of RGN, described in the main paper. 
    The RGN inference (top) and training (bottom) blocks correspond to (b) and (e) in Figure~\ref{fig:fig3} in the main paper, respectively.}
    \label{fig:supp_rgn_method_figure}
    \vspace{-1.0em}
\end{figure*}

\section{Details of RGN} \label{section:supp_rgn}
We illustrate the structure of RGN and the training/inference process in Figure~\ref{fig:supp_rgn_method_figure}. 
During training on each task, RGN $f^{RGN}_{T}$ learns to map the past principal rays from the previous RGN network $f^{RGN}_{T-1}$ and the current GT principal rays.
Such self-distillation training scheme is similar to how we incrementally train NeRF.
In inference time, we randomly select a number in $[0,1]$ and make a principal ray with RGN.
Then, using the camera intrinsic $K$, we randomly generate non-principal rays around principal rays.
As a result, we generate rays form a shape of cone emitted from the center of camera that covers the region of image plane.
This design helps reduce the burden of learning on RGN, at a small cost of accuracy.
Although we use a tiny network consisting of 3 layers, each with 16, 64, and 32 hidden units respectively, Figure~\ref{fig:supp_fig_ablation_RGNloss} demonstrates that even such a small network can remember the past principal rays.
This demonstrates the effectiveness of our RGN design.
\begin{figure}[t]
    \centering
    \includegraphics[width=\linewidth]{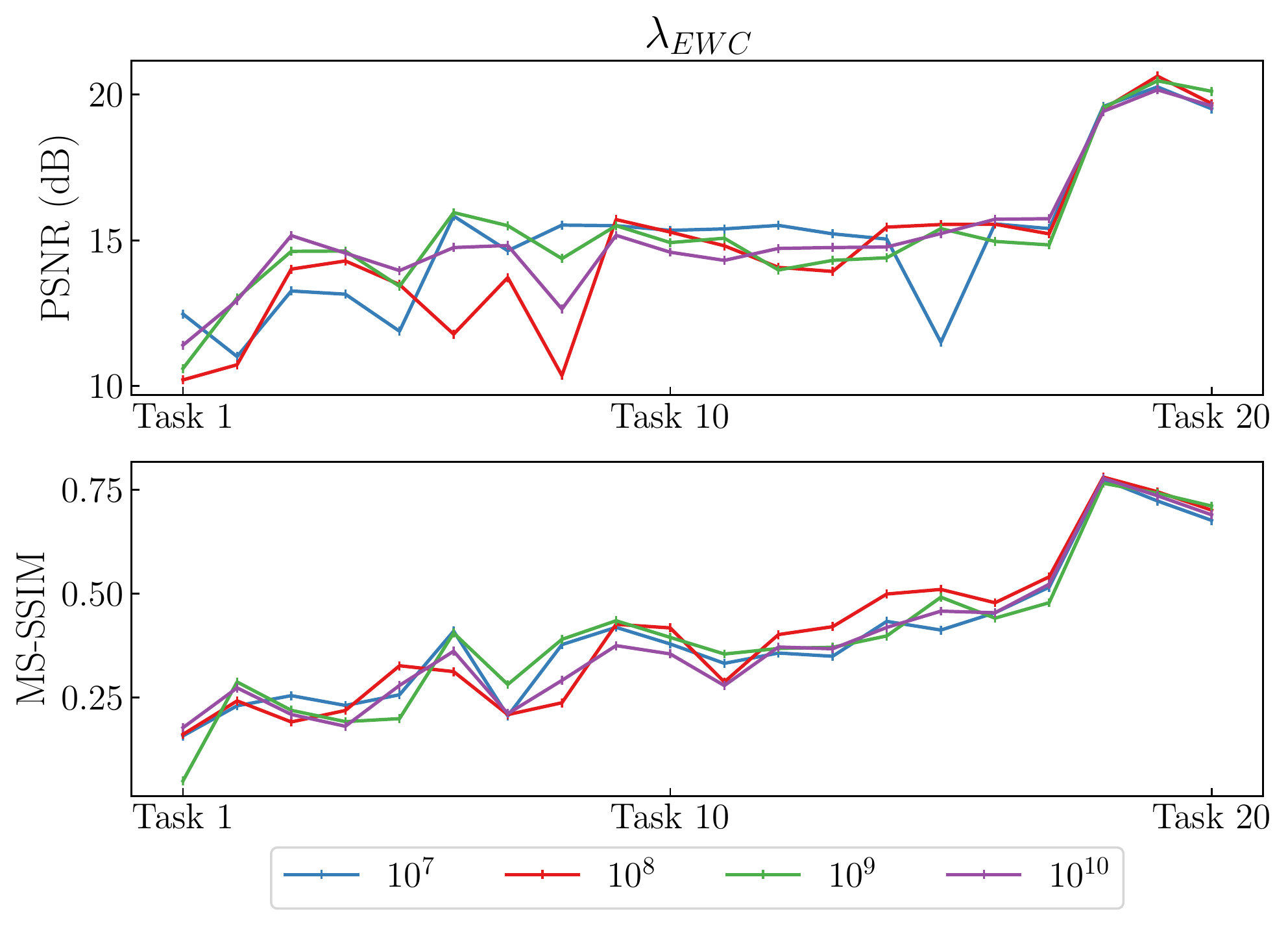}
    \vspace{-1.0em}
    \caption{\textbf{Ablation for EWC.} 
    We show the performance changes along the EWC weight, evaluated in Tanks and Temples/Barn}
    \label{fig:supp_fig_ablation_ewc}
    \vspace{-1.0em}
\end{figure}

The purpose of RGN is not to reproduce the rays of past tasks, but create the rays directed to the scene of interest near the past task cameras.
Therefore, errors under certain level are allowed when reconstructing the past cameras.
As shown in Figure~\ref{fig:supp_fig_ablation_RGN}, there are some accumulated errors in estimated camera poses. 
However, as shown in Figure~\ref{fig:ablation_RGN}, these generated and approximated rays are still effective in directing to the scene of interest near the past task cameras and thereby alleviating catastrophic forgetting.

\begin{figure}[t]
    \centering
    \includegraphics[width=\linewidth]{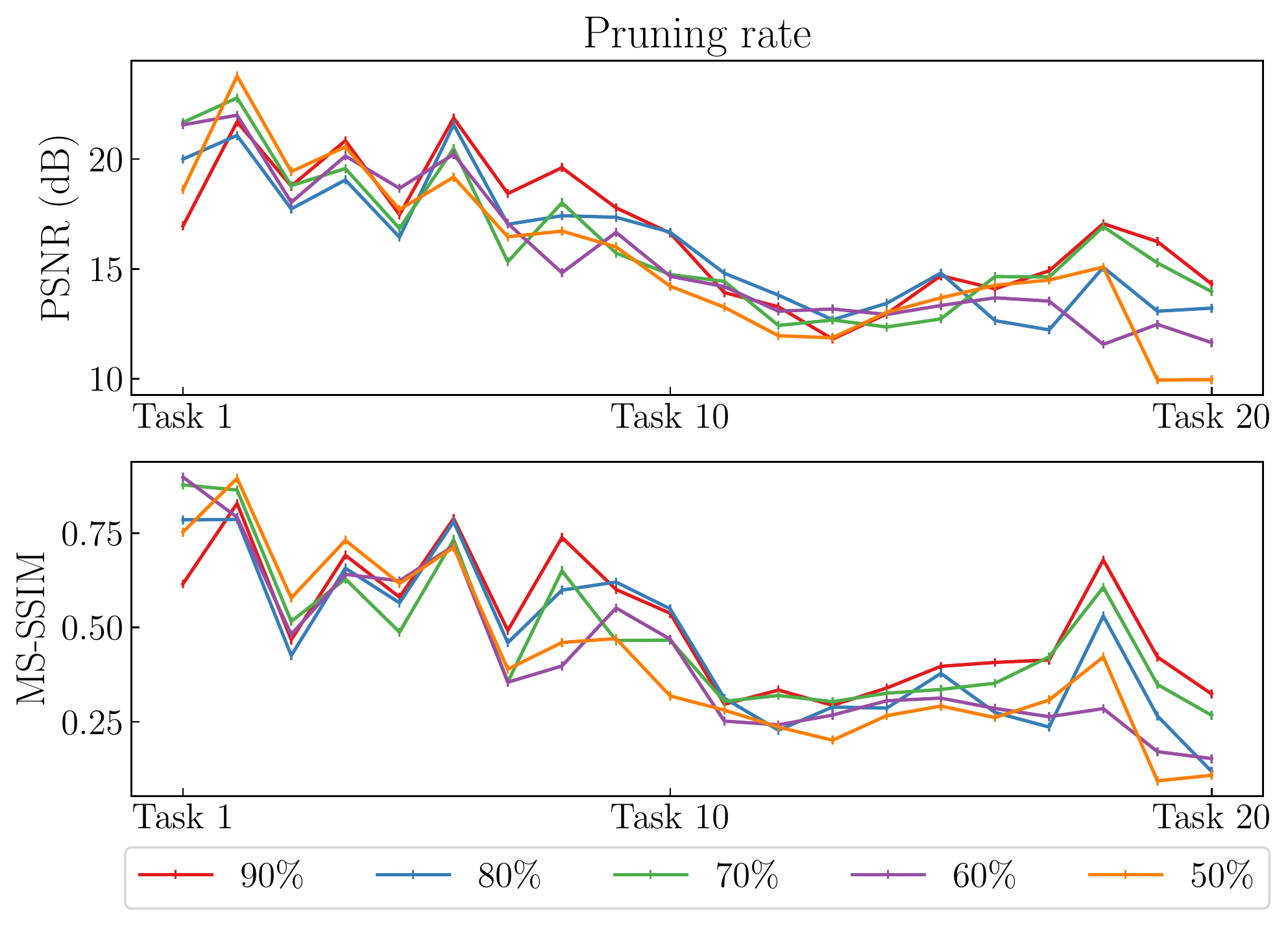}
    \vspace{-1.0em}
    \caption{\textbf{Ablation for PackNet.}
    We show the performance changes along the pruning rate in Packnet, evaluated in Tanks and Temples/Barn}
    \label{fig:supp_fig_ablation_packnet}
    \vspace{-1.0em}
\end{figure}

\section{Implementation details of other methods}
\label{section:other_methods}
In this section, we delineate the implementation details, especially hyperparameters, regarding other methods that are used for comparisons.
For fair comparisons, we choose the hyperparameters through ablation studies.

\subsection{Details of EWC} \label{section:supp_ewc_ablation}
We show the performance change in Figure~\ref{fig:supp_fig_ablation_ewc} as we vary the weight of parameter regularization loss term for EWC~\cite{kirkpatrick2017overcoming}.
Since the regularization loss is generally small, we multiply the loss by a large weight, as in the original EWC paper. 
Regardless of weight values, the figure exhibits severe catastrophic forgetting in EWC.
Throughout the experiments in this work, we use $10^8$ for the weight, making the regularization loss to be approximately 10\% of NeRF loss.

\subsection{Details of PackNet} \label{section:supp_packnet_ablation}
In PackNet~\cite{mallya2018packnet}, we repeat the process of training-pruning-retraining-freezing in order.
Depending on a pruning rate in this process, the performance of PackNet varies, as shown in Figure~\ref{fig:supp_fig_ablation_packnet}, 
In the figure, the following trends could be observed.
The higher the pruning rate, lower the performance on earlier tasks, but higher the performance on newer tasks.
On the other hand, the lower the pruning rate, the higher performance on earlier tasks, but the lower performance on newer tasks.
The observed trend is due to the fact that higher pruning rates lead to less parameters used for earlier tasks and more parameters reserved for newer tasks.
We set the pruning rate to 0.5 to follow the settings of the original paper while clearly showing the characteristics of PackNet in NeRF.

\input{sections/supp_table_ablation_replayLoss.tex}
\subsection{Details of iMAP} \label{section:supp_imap_ablation}
We perform an ablation study on a loss function used for remembering past tasks in the replay strategy, as in the proposed method (see Figure~\ref{fig:ablation_Loss} and Figure~\ref{fig:supp_ablation_loss_msssim}).
Table~\ref{tab:supp_ablation_replay_loss} shows the PSNR/MS-SSIM performance of iMAP trained with each loss function on TUM-RGBD/Seq0 data sequence.
L1 and Charbonnier, which provide significant improvement in our method, do not have much effect in the replay strategy. 
Compared to the default L2, there is a slight drop in PSNR and slight improvement in MS-SSIM. 
As the difference is not significant, we follow the original implementation and use L2 loss function for remembering past tasks in iMAP.
\clearpage


%% file: sections/supp_table_ablation_mcmp.tex
\renewcommand{\thetable}{S1}
\begin{table}[h!]
\centering
\caption{Ablation for batch size evaluated in Tanks and Temples/Barn}
\vspace{-0.8 em}
\label{tab:supp_ablation_mcmp}
\resizebox{\linewidth}{!}{
\large{
\begin{tabular}{|c|ccc|}
\hline
\multicolumn{1}{|c|}{\multirow{1}{*}{\small{PSNR / MS-SSIM}}} & $m_p=m_c/4$ & $m_p=m_c/2$ & $m_p=m_c$ \\ \hline
\multicolumn{1}{|c|}{$m_c=4096$} &
  21.92 / 0.821 &
  22.14 / 0.836 &
  22.30 / 0.835 \\ 
\multicolumn{1}{|c|}{$m_c=1024$} &
  21.72 / 0.805 &
  21.30 / 0.801 &
  21.82 / 0.812 \\ \hline
\end{tabular}}
}%
\vspace{-0.5em}
\end{table}

%% file: sections/supp_table_ablation_replayLoss.tex
\renewcommand{\thetable}{S2}
\begin{table}[h]
\centering
\caption{Ablation for the loss function used to train iMAP to remember past tasks. Evaluation is performed on TUM-RGBD/Seq0 data sequence.}
\vspace{-0.8 em}
\label{tab:supp_ablation_replay_loss}
\resizebox{\linewidth}{!}{
\large{
\begin{tabular}{|c|cc|}
\hline
\multicolumn{1}{|c|}{\multirow{1}{*}{(\textit{iMAP}) Loss for past tasks}} & PSNR (dB) & MS-SSIM \\ \hline
\multicolumn{1}{|c|}{L2 (iMAP*)} &
    \textbf{23.85} & 0.873 \\ 
\multicolumn{1}{|c|}{L1} &
    23.34 & 0.876 \\ 
\multicolumn{1}{|c|}{L1 + $\lambda_p$ scheduling} &
    23.70 & \textbf{0.886} \\ 
\multicolumn{1}{|c|}{Charbonnier} &
    23.40 & 0.876 \\ 
\multicolumn{1}{|c|}{Charbonnier + $\lambda_p$ scheduling} &
    23.70 & 0.884 \\ \hdashline[1pt/1.5pt]
\multicolumn{1}{|c|}{MEIL-NeRF (Ours)} &
    24.12 & 0.882 \\ \hline
\end{tabular}}
}%
\vspace{-0.5em}
\end{table}